\documentclass{article}

\usepackage{microtype}
\usepackage{graphicx}
\usepackage{subfigure}
\usepackage{booktabs} 

\usepackage{amsmath}
\usepackage{amssymb}

\usepackage{hyperref}

\makeatletter
\newenvironment{breakablealgorithm}
  {
   \begin{center}
     \refstepcounter{algorithm}
     \hrule height.8pt depth0pt \kern2pt
     \renewcommand{\caption}[2][\relax]{
       {\raggedright\textbf{\ALG@name~\thealgorithm} ##2\par}%
       \ifx\relax##1\relax 
         \addcontentsline{loa}{algorithm}{\protect\numberline{\thealgorithm}##2}%
       \else 
         \addcontentsline{loa}{algorithm}{\protect\numberline{\thealgorithm}##1}%
       \fi
       \kern2pt\hrule\kern2pt
     }
  }{
     \kern2pt\hrule\relax
   \end{center}
  }
\makeatother

\usepackage[accepted]{icml2019}

\icmltitlerunning{Model-based Deep Reinforcement Learning for Dynamic Portfolio Optimization}

\begin{document}

\twocolumn[
\icmltitle{Model-based Deep Reinforcement Learning for Dynamic Portfolio Optimization}



\icmlsetsymbol{equal}{*}

\begin{icmlauthorlist}
\icmlauthor{Pengqian Yu}{equal,neuri}
\icmlauthor{Joon Sern Lee}{equal,neuri}
\icmlauthor{Ilya Kulyatin}{neuri}
\icmlauthor{Zekun Shi}{neuri}
\icmlauthor{Sakyasingha Dasgupta}{neuri}
\end{icmlauthorlist}

\icmlaffiliation{neuri}{Neuri PTE LTD, One George Street, \#22-01, Singapore 049145}

\icmlcorrespondingauthor{Sakyasingha Dasgupta}{sakya@neuri.ai}

\icmlkeywords{Machine Learning, ICML}

\vskip 0.3in
]



\printAffiliationsAndNotice{\icmlEqualContribution} 

\begin{abstract}
{Dynamic portfolio optimization is the process of sequentially allocating wealth to a collection of assets in some consecutive trading periods, based on investors' return-risk profile. Automating this process with machine learning remains a challenging problem. Here, we design a deep reinforcement learning (RL) architecture with an autonomous trading agent such that, investment decisions and actions are made periodically, based on a global objective, with autonomy. In particular, without relying on a purely model-free RL agent, we train our trading agent using a novel RL architecture consisting of an infused prediction module (IPM), a generative adversarial data augmentation module (DAM) and a behavior cloning module (BCM). Our model-based approach works with both on-policy or off-policy RL algorithms. We further design the back-testing and execution engine which interact with the RL agent in real time. Using historical {\em real} financial market data, we simulate trading with practical constraints, and demonstrate that our proposed model is robust, profitable and risk-sensitive, as compared to baseline trading strategies and model-free RL agents from prior work.} 
\end{abstract}

\section{Introduction}

Reinforcement learning (RL) {consists of} an agent interacting with the environment, {in order to learn} an optimal policy by trial and error for sequential decision-making problems \cite{bertsekas2005dynamic, sutton2018reinforcement}. The past decade has witnessed the tremendous success of deep reinforcement learning (RL) in the fields of gaming, robotics and recommendation systems \cite{lillicrap2015continuous, silver2016mastering, mnih2015human, mnih2016asynchronous}. {However,} its applications in the financial domain have not been explored as thoroughly. 

Dynamic portfolio optimization remains one of the most challenging problems in the field of finance \cite{markowitz1959portfolio, haugen1990modern}. It is a sequential decision-making process of continuously reallocating funds into a number of different financial investment products, with the main aim to maximize return while constraining risk. Classical approaches to this problem include dynamic programming and convex optimization, which require {discrete actions} and thus suffer from the `curse of dimensionality' (e.g., \cite{cover91, li2014online, 6998078}). 

There have been efforts made to apply RL techniques to alleviate the dimensionality issue in the portfolio optimization problem \cite{moody2001learning, dempster2006automated, cumming2015investigation, jiang2017deep, deng2017deep, guo2018robust, liang2018adversarial}. The main idea is to train an RL agent that is rewarded if its investment decisions increase the logarithmic rate of return and is penalised otherwise. However, these RL algorithms have several drawbacks. In particular, the approaches in \cite{moody2001learning, dempster2006automated, cumming2015investigation, deng2017deep} only yield discrete single-asset trading signals. The multi-assets setting was studied in \cite{guo2018robust}, {however, the authors did not take transaction costs into consideration, thus limiting their practical usage}. In {recent study} \cite{jiang2017deep,liang2018adversarial}, transaction costs were considered but it did not address the challenge of having insufficient data in {financial markets} for the training of robust machine learning algorithms. {Moreover, the methods proposed in \cite{jiang2017deep,liang2018adversarial} directly apply a model-free RL algorithm that is sample inefficient and also doesn't account for the stability and risk issues caused by non-stationary financial market environment}.  In this paper, we propose a {novel model-based} RL approach, that takes into account practical trading restrictions such as transaction costs and order executions, to stably train an autonomous agent whose investment decisions are risk-averse yet profitable.  

{We highlight our three main contributions to realize a model-based RL algorithm for our problem setting}. Our first contribution is an infused prediction module (IPM), which incorporates the {prediction of expected future observations} into state-of-the-art RL algorithms. Our idea is inspired by some attempts to merge prediction methods with RL. {For example, RL has been successful in predicting the behavior of simple gaming environments} \cite{oh2015action}. In addition, {prediction based models have also been shown to improve the performance of RL agents} in distributing energy over a smart power grid \cite{marinescu2017prediction}. {In this paper}, we explore two prediction models; a nonlinear dynamic Boltzmann machine \cite{dasgupta2017nonlinear} and a variant of parallel WaveNet \cite{pmlr-v80-oord18a}. These models make use of historical prices of all assets in the portfolio to predict the future price movements of each asset, in a codependent manner. These predictions are then treated as additional features that can be used by the RL agent to improve its performance. Our experimental results show that using IPM provides significant performance improvements over baseline RL algorithms in terms of Sharpe ratio \cite{sharpe1966mutual}, Sortino ratio \cite{sortino1994performance}, maximum drawdown (MDD, see \cite{chekhlov2005drawdown}), value-at-risk (VaR, see \cite{artzner1999coherent}) and conditional value-at-risk (CVaR, see \cite{rockafellar2000optimization}). 

Our second contribution is a data augmentation module (DAM), which makes use of a generative adversarial network (GAN, e.g., \cite{goodfellow2014generative}) to generate synthetic market data. This module is well motivated by the fact that financial markets have limited data. To illustrate this, consider the case where new portfolio weights are decided by the agent on a daily basis. In such a scenario, which may not be uncommon, the size of the training set for a particular asset over the past 10 years is only around $2530,$ due to the fact that there are only about 253 trading days a year. Clearly, this is an extremely small dataset that may not be sufficient for training a robust RL agent. To overcome this difficulty, we train a recurrent GAN \cite{esteban2017real} using historical asset data to produce realistic multi-dimensional time series. Different from the objective {function} in \cite{goodfellow2014generative}, we explicitly consider the maximum mean discrepancy (MMD, see \cite{gretton2007kernel}) in the generator loss which further minimizes the distribution mismatch between real and generated data distributions. We show that DAM helps to reduce over-fitting and typically leads to {a} portfolio with less volatility. 

Our third contribution is a behavior cloning module (BCM), which provides one-step greedy {expert} demonstration to the RL agent. Our idea comes from the imitation learning {paradigm} (also called learning from demonstrations), with its most common form {being} behavior cloning, which learns a policy through {supervision provided by expert state-action pairs}. In particular, the agent receives examples of behavior from an expert and attempts to solve a task by mimicking the expert's behavior, e.g., \cite{bain1999framework, abbeel2004apprenticeship, ross2011reduction}. In RL, an agent attempts to maximize expected reward through interaction with the environment. Our proposed BCM combines aspects of conventional RL algorithms and supervised learning to solve complex tasks. This technique is similar in spirit to {the work in} \cite{nair2018overcoming}. The difference is that we create the expert behavior based on a one-step greedy strategy by solving an optimization problem that maximizes immediate rewards in the current time step. Additionally, we only update the actor with respect to its auxiliary behavior cloning loss in {an actor-critic algorithm} setting. We demonstrate that BCM can {prevent large changes in} portfolio weights and thus keep the volatility low, while also increasing returns in some cases.

To the best of our knowledge, this is the first work that leverages the deep RL state-of-art, and further extends it to a model-based setting and integrate it into the financial domain. Even though our proposed approach has been {rigorously} tested on the off-policy RL algorithm (in particular, the deep deterministic policy gradients (DDPG) algorithm \cite{lillicrap2015continuous}), these concepts can be easily extended to on-policy RL algorithms such as proximal policy optimization (PPO) \cite{schulman2017proximal} and trust region policy optimization \cite{schulman2015trust} algorithms. {We showcase the overall algorithm for model-based PPO for portfolio management and the corresponding results in the supplementary material.} Additionally, we also provide algorithms for differential risk sensitive deep RL for portfolio optimization in the supplementary material. For the rest of the main paper, our discussion will be centered around how our three contributions can improve the performance of the off-policy DDPG algorithm.

This paper is organized as follows: In Section \ref{sec2}, we review the deep RL literature, and formulate the portfolio {optimization} problem as a deep RL problem. We describe the structure of our automatic trading system in Section \ref{sec3}. Specifically, we provide details of the infused prediction module, data augmentation module and behavior cloning module in Section \ref{sec3.1} to Section \ref{sec3.3}. In Section \ref{sec4}, we report numerical experiments that serve to illustrate the {effectiveness of} methods described in this paper.  We conclude in Section \ref{sec5}. 

\section{Preliminaries and Problem Setup}\label{sec2}
In this section, we briefly review the literature of deep reinforcement learning and introduce the mathematical formulation of the dynamic portfolio optimization problem.

A Markov Decision Process (MDP) is defined as a 6-tuple $\langle T, \gamma, \mathcal{S}, \mathcal{A},  P, r\rangle$. Here, $T$ is the (possibly infinite) decision horizon; $\gamma\in(0,1]$ is the discount factor; $\mathcal{S}=\bigcup_t \mathcal{S}_t$ is the state space and $\mathcal{A}=\bigcup_t \mathcal{A}_t$ is the action space, both assumed to be finite dimensional and continuous; $P:\mathcal{S}\times\mathcal{A}\times\mathcal{S}\rightarrow[0,1]$ is the transition kernel and $r: \mathcal{S}\times\mathcal{A}\rightarrow\mathbb{R}$ is the reward function. Policy is a mapping $\mu: \mathcal{S}\rightarrow \mathcal{A}$, specifying the action to choose in a particular state.  At each time step $t\in\{1,\dots,T\}$, the agent in state $s_t\in\mathcal{S}_t$ takes an action $a_t=\mu(s_t)\in\mathcal{A}_t$, receives the reward $r_t$ and transits to the next state $s_{t+1}$ according to $P.$ The agent's objective is to maximize its expected return given the start distribution,
$
    J^\mu\triangleq\mathbb{E}_{s_t\sim P, a_t\sim \mu}[\sum_{t=1}^T \gamma^{t-1}r_t].
$
The state-action value function, or the Q value function, is defined as 
$
    Q^\mu(s_t,a_t)\triangleq\mathbb{E}_{s_{i>t}\sim P, a_{i>t}\sim \mu}[\sum_{i=t}^T\gamma^{(i-t-1)}r_i|s_t,a_t].
$

Deep deterministic policy {gradient} (DDPG) algorithm \cite{lillicrap2015continuous} is an off-policy model-free reinforcement learning algorithm for continuous control which utilize large function approximators such as {deep} neural networks. DDPG is an actor-critic method, which bridges the gap between policy gradient methods and value function approximation methods for RL. Intuitively, DDPG learns a state-action value function (critic) by minimizing the Bellman error, while simultaneously learning a policy (actor) by directly maximizing the estimated state-action value function with respect to the network parameters. 

In particular, DDPG maintains an actor function $\mu(s)$ with parameters $\theta^\mu$, a critic function $Q(s,a)$ with parameters $\theta^Q$, and a replay buffer $R$ as a set of tuples $(s_t, a_t, r_t, s_{t+1})$ for each experienced transition. DDPG alternates between running the policy to collect experiences (i.e. training roll-outs) and updating the parameters. In our implementation, training roll-outs were conducted with noise added to the policy network's parameter space to encourage exploration \cite{plappert2017parameter}. During each training step, DDPG samples a minibatch consisting of $N$ tuples from $R$ to update the actor and critic networks. DDPG minimizes the following loss $L$ w.r.t. $\theta^Q$ to update the critic, 
$
    L=\sum_{i=1}^N (r_i +\gamma Q(s_{i+1}, \mu(s_{i+1})) -
		Q(s_i, a_i | \theta^Q))^2/N.
$
The actor parameters $\theta^\mu$ are updated using the policy gradient
$
\nabla_{\theta^{\mu}} J^\mu =\sum_{i=1}^N
		\nabla_{a} Q(s, a | \theta^Q)|_{s = s_i, a = \mu(s_i)}\nabla_{\theta^\mu} \mu(s | \theta^\mu)|_{s =s_i}/N.
$
To stabilize learning, the Q value function is usually computed using a separate network (called the target network) whose weights are an exponential average over time of the critic network. This results in smoother target values.

Financial portfolio management is the process of constant redistribution of available funds to a set of financial assets. Our goal is to create a dynamic portfolio allocation system that periodically generates investment decisions and then act on these decisions autonomously. Following \cite{jiang2017deep}, we consider a portfolio of $m+1$ assets, including $m$ risky assets and $1$ risk-free asset (e.g., broker cash balance or U.S. treasury bond). We introduce such notation: given a matrix $\mathbf{g}$, we denote the $i^{\text{th}}$ row of $\mathbf{g}$ by $\mathbf{g}_{i,:}$, and the $j^{\text{th}}$ column by $\mathbf{g}_{:,j}$. We denote the closing, high and low price vectors of trading period $t$ as $\mathbf{p}_t$, $\mathbf{p}^{\text{h}}_t$ and $\mathbf{p}^{\text{l}}_t$ where $p_{i,t}$ is the closing price of the $i^{\text{th}}$ asset in the $t^{\text{th}}$ period. In this paper, we choose the first asset to be risk-free cash, i.e., $p_{0,t}=p_{0,t}^{\text{h}}=p_{0,t}^{\text{l}}=1$. We further define the price relative vector of the $t^{\text{th}}$ trading period as 
$
    \mathbf{u}_t\triangleq\mathbf{p}_t\oslash\mathbf{p}_{t-1}=(1,{p_{1,t}}/{p_{1,t-1}},\dots,{p_{m,t}}/{p_{m,t-1}})^\top
$
where $\oslash$ denotes the element-wise division. In addition,  we let $h_{i,t}\triangleq(p_{i,t}-p_{i,{t-1}})/p_{i,{t-1}}$ denote the percentage change of closing price at time $t$ for asset $i$, the space associated with its vector form $\mathbf{h}_{:,t}$ ($\mathbf{h}_{i,:}$) as $\mathcal{H}_{:,t}\subset\mathbb{R}^m$ ($\mathcal{H}_{i,:}\subset \mathbb{R}^{k_1}$) where $k_1$ is the time embedding of prediction model. We define $\mathbf{w}_{t-1}$ as the portfolio weight vector at the beginning of trading period $t$ where its $i^{\text{th}}$ element $w_{i,t-1}$ represents the proportion of asset $i$ in the portfolio after capital reallocation and $\sum_{i=0}^{m}w_{i,t}=1$ for all $t$. We initialize our portfolio with $\mathbf{w}_0=(1,0,\dots,0)^\top$.  Due to price movements in the market, at the end of the same period, the weights evolve according to
$\mathbf{w}_t^{'}={(\mathbf{u}_t\odot\mathbf{w}_{t-1})}/{(\mathbf{u}_t\cdot\mathbf{w}_{t-1})},$
where $\odot$ is the element-wise multiplication. Our goal at the end of period $t$ is to reallocate portfolio vector from $\mathbf{w}^{'}_t$ to $\mathbf{w}_t$ by selling and buying
relevant assets. Paying all commission fees, this reallocation action shrinks the portfolio value by a factor $\bar{c}_t\triangleq c\sum_{i=1}^m|w_{i,t}^{'}-w_{i,t}|$ where $c$ is the transaction fees for purchasing and selling. In particular, we let $\rho_{t-1}$ denote the portfolio value at the beginning of period $t$ and $\rho^{'}_{t}$ at the end. We then have $\rho_t=\bar{c}_t\rho^{'}_{t}.$ The immediate reward is the logarithmic rate of return defined by
$
    r_t\triangleq\ln({\rho_t}/{\rho_{t-1}})=\ln ({\bar{c}_t\rho^{'}_t}/{\rho_{t-1}})=\ln (\bar{c}_t\mathbf{u}_t\cdot\mathbf{w}_{t-1}).
$

We define the normalized close price matrix at time $t$ by $\mathbf{P}_t\triangleq[\mathbf{p}_{t-k_2+1}\oslash\mathbf{p}_t|\mathbf{p}_{t-k_2+2}\oslash\mathbf{p}_t|\cdots|\mathbf{p}_{t-1}\oslash\mathbf{p}_{t}|\mathbf{1}]$ where $\mathbf{1}\triangleq(1,1,\dots,1)^\top$ and $k_2$ is the time embedding. The normalized high price matrix $\mathbf{P}_t^{\text{h}}$ is defined by 
$\mathbf{P}^{\text{h}}_t\triangleq[\mathbf{p}^{\text{h}}_{t-k_2+1}\oslash\mathbf{p}_t|\mathbf{p}^{\text{h}}_{t-k_2+2}\oslash\mathbf{p}_t|\cdots|\mathbf{p}^{\text{h}}_{t-1}\oslash\mathbf{p}_{t}|\mathbf{p}^{\text{h}}_{t}\oslash\mathbf{p}_{t}]$, and low price matrix $\mathbf{P}_t^{\text{l}}$ can be defined similarly. 
We further define the price tensor as $\mathbf{Y}_t\triangleq[\mathbf{P}_t\,\,\mathbf{P}_t^{\text{h}}\,\,\mathbf{P}_t^{\text{l}}]$.  Our objective is to design a RL agent that observes the state $s_t\triangleq(\mathbf{Y}_t, \mathbf{w}_{t-1})$ and takes a sequence of actions (portfolio weights) over the time $a_t=\mathbf{w}_t$ such that the final portfolio value 
$\rho_{T}=\rho_0\mathbb{E}_{s_t\sim P, a_t\sim \mu}[\exp(\sum_{t=1}^{T}\gamma^{t-1}r_t)]$
is maximized.

\section{System Design and Data}\label{sec3}
In this section, we discuss the detailed design of our proposed RL based automatic trading system.

\begin{figure}[ht]
\begin{center}
\centerline{\includegraphics[width=2.5in]{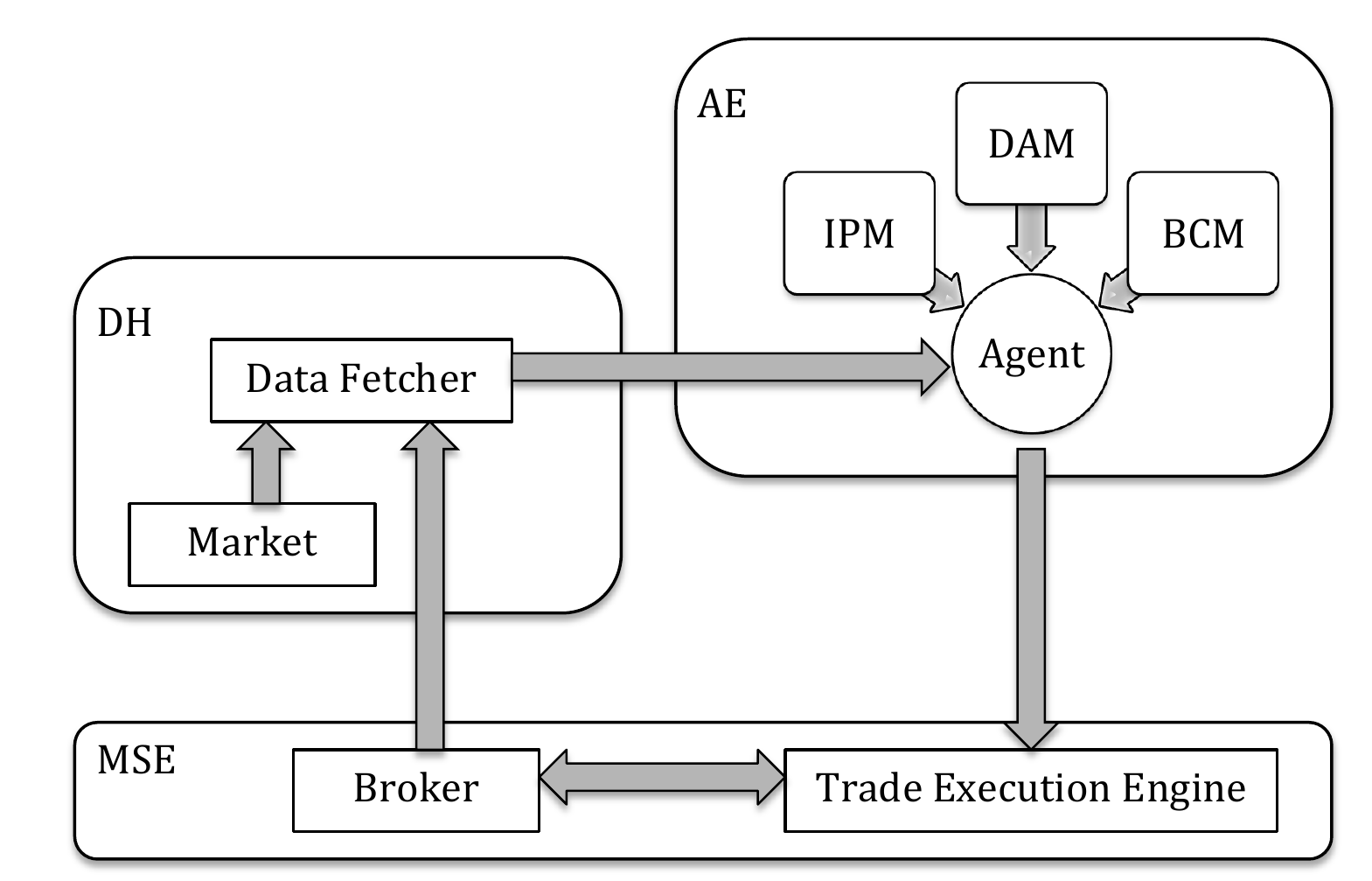}}
\caption{Trading framework.}
\label{system}
\end{center}
\vskip -0.3in
\end{figure}

The trading framework referenced in this paper is represented in Figure \ref{system} and is a modular system composed of a data handler (DH), an algorithm engine (AE) and a market simulation engine (MSE). The DH retrieves market data and deals with the required data transformations. It is designed for continuous data ingestion in order to provide the AE with the required set of information. The AE is a collection of models containing RL agents and environment specifications. We refer the readers to Algorithm \ref{algo1} in the supplementary material for further details. The MSE is an online event-driven module that provides feedback of executed trades, which can eventually be used by the AE to compute rewards. In addition, it also executes investment decisions made by the AE. The strategy applied in this study is an asset allocation system that rebalances the available capital between the provided set of assets and a cash asset on a daily frequency.

The data used in this paper is a mix of U.S. equities\footnote{We use data from Refinitiv DataScope, with experiments carried out on the following U.S. Equities: Costco Wholesale Corporation, Cisco Systems, Ford Motors, Goldman Sachs, American International Group and Caterpillar. The selection is qualitative with a balanced mix of volatile and less volatile stocks. An additional asset is cash (in U.S. dollars), representing the amount of capital not invested in any other asset. Furthermore, a generic market variable (S\&P 500 index) has been added as an additional feature. The data is shared in supplementary files, which will be made available publicly later.} on tick level (trade by trade) aggregated to form open-high-low-close (OHLC) bars on an hourly frequency\footnote{We execute orders hourly, while the agent produces decisions daily.}.

In the financial domain, it is common to use benchmark strategies to evaluate the relative profitability and risk profile of the tested strategies. A common benchmark strategy is the constantly rebalanced portfolio (CRP), where at each period the portfolio is rebalanced to the initial wealth distribution among the $m+1$ assets including the cash. This strategy can be seen as using the mean-reverting nature of stock prices, as it sells those that gained value while buying more of those losing value. In \cite{cover91} it is shown that such a strategy is asymptotically the best for stationary stochastic processes such as stock prices, offering exponential returns and is, therefore, an optimistic benchmark to compare against. Transaction fees $c$ have been fixed at a conservative level of 20 basis points\footnote{One basis point is equivalent to 0.01\%.} and, given the use of market orders, an additional 50 basis points slippage\footnote{Slippage is defined as the relative deviation of the price at which the orders get executed and the price at which the agent produced the reallocation action. This conservative level of slippage is due to the lack of equity volume data, as during less liquid trading periods the agent market orders might influence the market price and get a worse trade execution than expected in a simulated environment.} is applied.

Performance monitoring is based on a set of evaluation measures such as Sharpe and Sortino ratios, value-at-risk (VaR), conditional value-at-risk (CVaR), maximum drawdown (MDD), annualized volatility and annualized returns. Let $Y$ denote a bounded random variable. The Sharpe ratio of $Y$ is defined as
$
    \text{SR}\triangleq{\mathbb{E}[Y]}/{\sqrt{\text{var}[Y]}}.
$
Sharpe ratio, representing the reward per unit of risk, has been recognized not to be desirable since it is a symmetric measure of risk and, hence, penalizes the low-cost events. Sortino ratio, VaR and CVaR are risk measures which gained popularity for taking into consideration only the unfavorable part of the return distribution, or, equivalently, unwanted high cost. Sortino ratio is defined similarly to Sharpe ratio,  though replacing the standard deviation $\sqrt{\text{var}[Y]}$ with the downside deviation. The $\text{VaR}_{\alpha}$ with level $\alpha\in(0,1)$ of $Y$ is the $(1-\alpha)$-quantile of $Y$, and $\text{CVaR}_{\alpha}$ at level $\alpha$ is the  expected return of $Y$ in the worst $(1-\alpha)$ of cases.

\subsection{Network Architecture}
In this subsection, we provide a qualitative description of our proposed network before delving into the mathematical details in the subsequent subsections.

\begin{figure}[ht]
\vskip -0.2in
\begin{center}
\centerline{\includegraphics[width=3.2in]{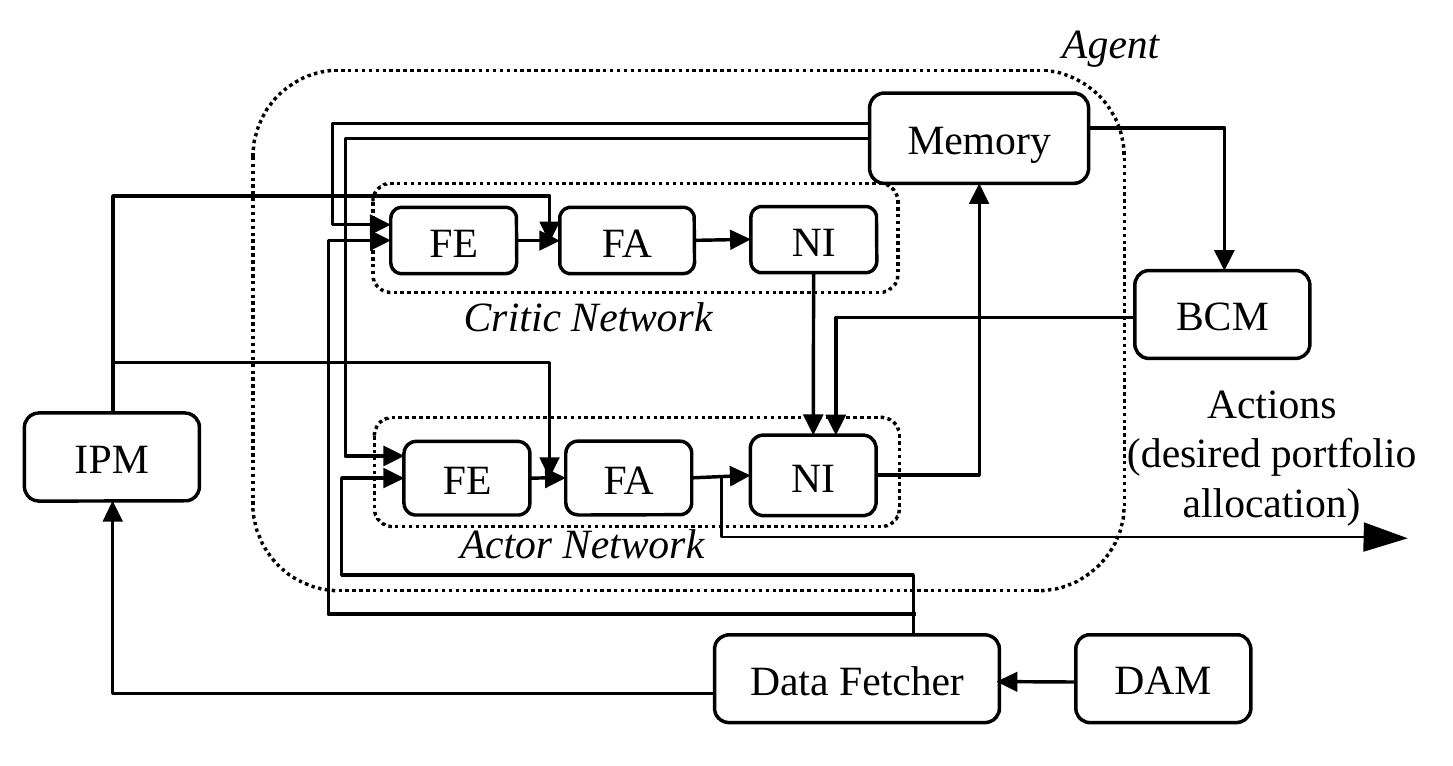}}
\caption{Agent architecture.}
\label{agent}
\end{center}
\vskip -0.3in
\end{figure}

Figure \ref{agent} shows how we integrate the IPM, BCM and DAM into our network architecture. In particular, the DAM is used to append generated data to each training episode. The data fetcher fetches data from the augmented data in a step-wise fashion. This data is then fed to both the IPM and agent, which have multiple neural networks. 

To illustrate our proposed off-policy version of dynamic portfolio optimization algorithm, we adapt the actor-critic style DDPG algorithm \cite{lillicrap2015continuous}. In this setting, at least two networks (one for the actor and one for the critic) are required in the agent as shown in Figure \ref{agent}. Furthermore, our implementation utilizes both target networks \cite{mnih2015human} and parameter noise exploration \cite{plappert2017parameter}, which in itself necessitates two additional networks for the actor. Our agent, comprising of six separate networks (four networks for the actor and two networks for the critic), is described in Algorithm \ref{algo1} of supplementary material Section A. 

We next discuss how the agent is trained and tested. For each episode during training, we select an episode of data by selecting a random trading date that satisfies our desired episode length. The training data of each episode is then augmented by synthetic data generated by the DAM. At each time step of the episode, market data, i.e., data corresponding to the current time step of interest, fetched via the data fetcher is used to train the IPM which produces predictions of price percentage changes in the next time step. At the same time, the agent's networks are updated by making use of data sampled from the memory (also referred to as the replay buffer) via the prioritized experience replay \cite{schaul2015prioritized}. Once the agent's networks are updated, an action, i.e., desired portfolio allocation in the next time step, can be obtained from the actor network that has been perturbed with the parameter noise. The MSE then executes this noisy action and the agent moves to the next state. The corresponding state, action, rewards, next state and computed one-step greedy action, produced by the BCM, are stored in the memory. This process is repeated for each step in the episode and for all episodes. During the testing period, the IPM continues to be updated at each time step while the agent is frozen in that it is no longer trained and actions are obtained from the actual actor network, i.e., the actor network without parameter noise.

As shown in Figure \ref{agent}, actor and critic networks in the agent consist of feature extraction (FE), feature analysis (FA) and network improvement (NI) components. The FE layers aim to extract features given the current price tensor $\mathbf{Y}_t$. In our experiments, we have $8$ assets including cash and a time embedding $k_2=10$. This essentially means that we have a price tensor of the shape $3\times8\times10$ if using the channels first convention.  The FE layers can be either LSTM-based recurrent neural networks (RNN, see \cite{hochreiter1997long}) or convolutional neural networks (CNN, see \cite{krizhevsky2012imagenet}). We find that the former typically yields better performance. Thus, the price tensor is reshaped to a $24\times10$ tensor prior to being fed to the LSTM-based FE network. 

The outputs at each time step of the LSTM-based FE network are concatenated together into a single vector. Next, the previous actions, $\mathbf{w}_{t-1}$, is concatenated to this vector. Finally, it is further concatenated with a one-step predicted price percentage change vector produced by the IPM and a market index performance indicator (i.e., the price ratio of a market index such as the S\&P 500 index). The resulting vector is then passed to a series of dense layers (i.e., multilayer perceptrons), which we refer to as the feature analysis (FA) component. We remark that the network in \cite{jiang2017deep} does not have dense layers, which may not account for non-linear relationships across various assets.

Finally, we have a network improvement (NI) component for each network which specifies how the network is updated. Specifically, NI synchronizes the learning rates between the actor and the critic, which preserves training stability by ensuring the actor is updated at a slower rate than the critic \cite{bhatnagar2009natural}. It is also important to note that the actor network's NI component receives gradients from the BCM, which makes use of one-step greedy actions to provide supervised updates to the actor network to reduce portfolio volatility. 

\subsection{Infused Prediction Module}\label{sec3.1}
Here, for the IPM, we implemented and evaluated two different multivariate prediction models, trained in an online manner, differing in their computational complexity. As the most time efficient model, we implemented the nonlinear dynamic Boltzmann machine \cite{dasgupta2017nonlinear} (NDyBM) that predicts the future price of each asset conditioned on the history of all assets in the portfolio. As NDyBM does not require backpropagation through time, it has a parameter update time complexity of $\mathcal{O}(1)$. This makes it very suitable for fast computation in online time-series prediction scenarios. However, NDyBM assumes that the inputs come from an underlying Gaussian distribution. In order to make IPM more generic, we also implemented a novel prediction model using dilated convolution layers inspired by the WaveNet architecture \cite{pmlr-v80-oord18a}. As there was no significant difference in predictive performance noticed between the two models, in the rest of the paper we provide results with the faster NDyBM based IPM module. However, details of our WaveNet inspired architecture can be seen in supplementary section D. 

We use the state-space augmentation technique, and construct the augmented state-space $\widetilde{\mathcal{S}}\triangleq\mathcal{S}_t\times\mathcal{H}_{:,t+1}\times\mathcal{H}_{:,t+1}\times\mathcal{H}_{:,t+1}$: each state is now a pair $\tilde{s}_t\triangleq(s_t,\mathbf{x}_{t+1})$, where $s_t\in\mathcal{S}_t$ is the original state, and $\mathbf{x}_{t+1}\triangleq(\mathbf{h}_{:,t+1}, \mathbf{h}^{\text{h}}_{:,t+1}, \mathbf{h}^{\text{l}}_{:,t+1})\in\mathcal{X}\subset\mathbb{R}^{m\times 3}$ where $\mathbf{h}_{:,t+1}, \mathbf{h}^{\text{h}}_{:,t+1}, \mathbf{h}^{\text{l}}_{:,t+1}\in\mathcal{H}_{:,t+1}$ is the predicted future close, high and low asset percentage price change tensor. 

The NDyBM can be seen as an unfolded Gaussian Boltzmann machine for an infinite time horizon i.e. $T \rightarrow \infty$ history, that generalizes a standard vector auto-regressive model with eligibility traces and nonlinear transformation of historical data  \cite{dasgupta2017nonlinear}. It represents the conditional probability density of 
$\mathbf{x}^{[t]}$ given $\mathbf{x}^{[:t-1]}$ as, $ p(\mathbf{x}^{[t]} | \mathbf{x}^{[:t-1]})
 = \prod_{j=1}^N p_j(x_j^{[t]} | \mathbf{x}^{[:t-1]})$\footnote{For mathematical convenience, $\mathbf{x}_t$ and $\mathbf{x}^{[t]}$ are used interchangeably. }. Where, each factor of the right-hand side denotes the conditional probability density of $x_j^{[t]}$ given
$\mathbf{x}^{[:t-1]}$ for $j=1,\ldots,N$. Where, $N= m\times 3$ are the number of units in the NDyBM. Here, $x_j^{[t]}$ is considered to have a Gaussian distribution for each $j$:\\$
 p_j(x_j^{[t]} | \mathbf{x}^{[:t-1]}) = \frac{1}{\sqrt{2\,\pi\,\sigma_j^2}}\exp\Big(-\frac{\big(x_j^{[t]}-\mu_j^{[t]}\big)^2}{2\sigma_j^2}\Big)$. \\ Here,  
$\boldsymbol{\mu}\triangleq(\mu_j)_{j=1,\ldots,N}$ is the vector of expected values of the $j$-th unit at time $t$ given the history up to $t-1$ patterns. It is represented as: \\
$\boldsymbol{\mu}^{[t]}= \mathbf{b} + \sum_{\delta=1}^{d-1} \mathbf{F}^{[\delta]}\mathbf{x}^{[t-\delta]} + \sum_{k=1}^K \mathbf{G}_k \, \boldsymbol{\alpha}_k^{[t-1]}$.\\
Where, $\mathbf{b}\triangleq(b_j)_{j=1,\ldots,N}$ is a bias vector,
$\boldsymbol{\alpha}_k^{[t-1]}\triangleq(\alpha_{j,k}^{[t-1]})_{j=1,\ldots,N}$ are $K$ eligibility trace vectors, $d$ is the time-delay between connections $(i,j)$ and \\
$\mathbf{F}^{[\delta]}\triangleq(\tilde f_{i,j})_{(i,j)\in\{1,\ldots,N\}^2}$ for $0<\delta<d$, $\mathbf{G}_k\triangleq(g_{i,j,k})_{(i,j)\in\{1,\ldots,N\}^2}$ for $k=1,\ldots,K$ are $N\times N$ weight matrices. The eligibility trace can be updated recursively as, $\alpha_{i,j,k}^{[t]} = \lambda_k \, \alpha_{i,j,k}^{[t-1]} + x_i^{[t-d_{i,j}+1]}$. Here, $\lambda_k$ is a fixed decay rate factor defined for each of the $k$ column vectors. Additionally, the bias parameter vector $\mathbf{b}$, is updated at each time using a RNN layer. This RNN layer computes a nonlinear feature map of the past time series. Where in, the output weights from the RNN to the bias layer along with other NDyBM parameters (bias, weight matrices and variance), are updated online using a stochastic gradient method. 

Following \cite{dasgupta2017nonlinear}\footnote{Details of the learning rule and derivation of the model as per the original paper. Algorithm steps and hyper-parameter settings are provided in supplementary.}, the NDyBM is trained to predict the next time-step close, high and low percentage change for each asset conditioned on the history of all other assets, such that the log-likelihood of each time-series is maximised. As such NDyBM parameters are updated at each step $t$, following the gradient of the conditional probability density of $\mathbf{x}^{[t]}$: $\nabla_\theta \log p(\mathbf{x}^{[t]}|\mathbf{x}^{[-\infty,t-1]}) = \sum_{i=1}^N \nabla_\theta \log p_i(x_i^{[t]}|\mathbf{x}^{[-\infty,t-1]})$.

In the spirit of providing the agent with additional market signals and removing non-Markovian dynamics in the model, along with the prediction, we further augment the state space with a market index performance indicator. The state space now has the form $\tilde{s}_{t}\triangleq(s_t,\mathbf{x}_{t+1},I_t)$ where,  $I_t\in\mathbb{R}_{+}$ is the market index performance indicator for time step $t$.

\subsection{Data Augmentation Module}\label{sec3.2}
The limited historical financial data prevents us from scaling up the deep RL agent. To mitigate this issue, we augment the data set via the recurrent generative adversarial networks (RGAN) framework \cite{esteban2017real} and generate synthetic time series. The RGAN follows the architecture of a regular GAN, with both the generator and the discriminator substituted by recurrent neural networks. Specifically, we generate percentage change of closing price for each asset separately at higher frequency than what the RL agent uses. We then downsample the generated series to obtain the synthetic series of high, low, close (HLC) triplet vector $\mathbf{x}_t$. This approach avoids the non-stationarity in market price dynamics by generating a stationary percentage change series instead, and the generated HLC triplet is guaranteed to maintain their relationship (i.e., generated highs are higher than generated lows).


We assume implicitly that the $i^{\text{th}}$ assets' percentage change vector $\mathbf{h}_{i,:}$ follows a distribution: $\mathbf{h}_{i,:}\sim p_{\text{data}}^i(\mathbf{h}_{i,:})$. Let $H$ be the hidden dimension. Our goal is to find a parameterized function $G^i_\psi$ such that given a noise prior $p_z(\mathbf{z}), \mathbf{z}\in\mathbb{R}^{k_1\times H}$ the generated distribution $p^i_g(G^i_\psi(\mathbf{z}))$ are \textit{empirically similar} to the data distribution $p_{\text{data}}^i(\mathbf{h}_{i,:})$. Formally, given two batches of observations  $\{G^i_\psi(\mathbf{z})^{(j)}\}^b_{j=1}, \{\mathbf{h}_{i,:}^{(j)}\}^b_{j=1}$ from distributions $p^i_g$ and $p_{\text{data}}^i$ where $b$ is the batch size, we want $p^i_g\approx p^i_{\text{data}}$ under certain similarity measure of distributions.

One suitable choice of such similarity measure is the maximum mean discrepancy (MMD) \cite{gretton2012}. Following \cite{yujia2015gmmn}, we can show (see supplementary material) that with RBF kernel, minimizing $\widehat{\text{MMD}}^2_b$, which is the biased estimator of squared MMD results in matching all moments between the two distribution $p_a, p_b$. 

As discussed in previous works \cite{arjovsky17a, arjovsky2017towards}, vanilla GANs suffer from the problem that discriminator becomes perfect when the real and the generated probabilities have disjoint supports (which is often the case under the hypothesis that real-world data lies in low dimensional manifolds). This could lead to generator gradient vanishing, making the training difficult. Furthermore, the generator can suffer from mode collapse issue where it succeeds in tricking the discriminator but the generated samples have low variation. 

In our RGAN architecture, we model each asset $i$ separately by a pair of parameterized function $D^i_\phi, G^i_\psi$, and we use $\widehat{\text{MMD}}^2$, the unbiased estimator of squared MMD between $p^i_g$ and $p^i_{\text{data}}$, as a regularizer for the generator $G^i_\psi$, such that the generator not only tries to 'trick' the discriminator into classifying its output as coming from $p^i_{\text{data}}$, but also tries to match $p^i_g$ with $p^i_{\text{data}}$ in all moments. This alleviates the aforementioned issues in vanilla GAN: firstly $\widehat{\text{MMD}}^2$ is defined even when distributions have disjoint supports, and secondly gradients provided by $\widehat{\text{MMD}}^2$ are not dependent on the discriminator but only on the real data. Specifically, both the discriminator $D^i_\phi$ and the generator $G^i_\psi$ is trained with gradient descent method. The discriminator objective is: \\
$ \max_\phi\,\frac{1}{b} \sum_{j=1}^b \left[ \log D^i_\phi(\mathbf{h}_i^{(j)}) + \log\left(1-D^i_\phi ( G^i_\psi (\mathbf{z}^{(j)}) )\right) \right]$, 

and the generator objective is: \\ 
$ \min_\psi\,\frac{1}{b} \sum_{j=1}^b \left[ \log \left( 1 - D^i_\phi ( G^i_\psi (\mathbf{z}^{(j)}) )  \right)\right] + \zeta\widehat{\text{MMD}}^2$

given a batch of $b$ samples $\{\mathbf{z}^{(j)}\}^b_{j=1}$ drawn independently from a diagonal Gaussian noise prior $p_z(\mathbf{z})$, and a batch of $b$ samples $\{\mathbf{h}_i^{(j)}\}^b_{j=1}$ drawn from the data distribution $p_{\text{data}}^i$ of the $i^{\text{th}}$ asset. \newline
In order to select the bandwidth parameter $\sigma$ in the RBF kernel, we set it to the median pairwise distance between the joint data. Both discriminator and generator networks are LSTMs \cite{hochreiter1997long}. \\
Although generator directly minimises estimated $\widehat{\text{MMD}}^2$, we go one step further to validate the RGAN by conducting Kolmogorov-Smirnov (KS) test. Our results show that the RGAN is indeed generating data representative of the true underlying distribution. More details of the KS test can be found in supplementary material.

\subsection{Behavior Cloning Module}\label{sec3.3}
In finance, some investors may favor a portfolio with lower volatility over the investment horizon. To achieve this, we propose a novel method of behavior cloning, with the primary purpose of reducing portfolio volatility while maintaining reward to risk ratios. We remark that one can broadly dichotomize  imitation learning into a passive collection of demonstrations (behavioral cloning) versus an active collection of demonstrations. The former setting \cite{abbeel2004apprenticeship, ross2011reduction} assumes that demonstrations are collected a priori and the goal of imitation learning is to find a policy that mimics the demonstrations. The latter setting \cite{daume2009search, sun2017deeply} assumes an interactive expert that provides demonstrations in response to actions taken by the current policy. Our proposed BCM in this definition is an active imitation learning algorithm.  
In particular, for every step that the agent takes during training, we calculate the one-step greedy action in hindsight. This one-step greedy action is computed by solving an optimization problem, given the next time period's price ratios, current portfolio distribution and transaction costs. The objective function is to maximize returns in the current time step, which is why the computed action is referred to as the one-step greedy action. For time step $t$, the objective function is as follows: 
\begin{equation}\label{expert}
\begin{aligned}
    \max_{\mathbf{w}_{t}}&\quad\mathbf{u}_{t}\cdot\mathbf{w}_{t}-c\textstyle\sum_{i=1}^m|{w}_{i,t}-{w}_{i,t-1}|\\
    \text{s.t.}&\quad \textstyle\sum_{i=0}^m w_{i,t}=1,\quad 0\leq w_{i,t}\leq1,\,\forall i.
\end{aligned}
\end{equation}
Solving the above optimization problem for $\mathbf{w}_t$ yields an optimal expert greedy action denoted by $\bar{a}_t$. This one-step greedy action is then stored in the replay buffer together with the corresponding $(s_t, a_t, r_t, s_{t+1})$ pair. In each training iteration of the actor-critic algorithm, a mini-batch of $\{(s_i, a_i, r_i, s_{i+1}, \bar{a}_i)\}_{i=1}^N$ is sampled from the replay buffer. Using the states $s_i$ that were sampled from the replay buffer, the actor's corresponding actions are computed and the log-loss between the actor's actions and the one-step greedy actions $\bar{a}_i$ is calculated: 
\vspace{-0.06in}
\begin{equation}\label{expert_loss}
    \begin{aligned}
        &\bar{L}^\mu=-1\times\\&\frac{\sum_{i=1}^N\sum_{j=0}^m \bar{a}_{i,j}\log(\mu(s_i))_j+(1-\bar{a}_{i,j})\log(1-(\mu(s_i))_j)}{N(m+1)}.
    \end{aligned}
\end{equation}

Gradients of the log-loss with respect to the actor network $\nabla_{\theta^\mu}\bar{L}^\mu$ can then be calculated and used to perturb the weights of the actor slightly.  Using this
loss directly prevents the learned policy from improving significantly beyond the demonstration policy, as the actor is always tied back to the demonstrations. To achieve this, a factor $\lambda$ is used to discount the gradients such that the actor network is only slightly perturbed towards the one-step greedy action, thus maintaining the stability of the underlying RL algorithm. Following this, the typical DDPG algorithm as described above is executed to train the agent.

\begin{table*}[t]
\caption{Performances for different models (last column represents IPM+DAM+BCM model. ann. and accnt. stand for annualized and account, respectively).}
\label{performance}
\begin{center}
\begin{small}
\begin{tabular}{|c|c|c|c|c|c|c|c|c|c|}
\hline

                      & CRP    & Baseline & IPM              & DAM     & BCM     & IPM+DAM & IPM+BCM & DAM+BCM          & All      \\ \hline
Final accnt. value   & 574859 & 570482   & \textbf{586150}  & 571853  & 571578  & 575430  & 577293  & 571417           & 580899           \\ \hline
Ann. return     & 7.25\% & 7.14\%   & \textbf{8.64\%}  & 7.26\%  & 7.24\%  & 7.60\%  & 7.77\%  & 7.22\%           & 8.09\%           \\ \hline
Ann. volatility & \textbf{12.65\%}& 12.79\%  & 14.14\%          & 12.80\% & 12.79\% & 12.84\% & 12.81\% & {12.77\%} & {12.77\%} \\ \hline
Sharpe ratio          & 0.57    & 0.56     & 0.61             & 0.57    & 0.57    & 0.59    & 0.61    & 0.57             & \textbf{0.63}    \\ \hline
Sortino ratio         & 0.80   & 0.78     & 0.87             & 0.79    & 0.79    & 0.83    & 0.85    & 0.79             & \textbf{0.89}    \\ \hline
$\text{VaR}_{0.95}$                   & \textbf{1.27\%} & 1.30\%   & 1.41\%           & 1.30\%  & 1.30\%  & 1.29\%  & 1.28\%  & 1.30\%           & \textbf{1.27\%}  \\ \hline
$\text{CVaR}_{0.95}$                  & \textbf{1.91\%} & 1.93\%   & 2.11\%           & 1.94    & 1.93\%  & 1.93\%  & 1.92\%  & 1.93\%           & \textbf{1.91\%}  \\ \hline
MDD                   & 13.10\% & 13.80\%  & \textbf{12.20\%} & 13.70\% & 13.70\% & 12.70\% & 12.60\% & 13.70\%          & 12.40\%          \\ \hline
\end{tabular}
\end{small}
\end{center}
\vskip -0.1in
\end{table*}
\section{Experiments}\label{sec4}
All experiments\footnote{Algorithm of our model-based DDPG agent is detailed in supplementary material Section A.} were conducted using a backtesting environment, and the assets used are as described in Section \ref{sec3}. The time period spanning from 1 Jan 2005 to 31 Dec 2016 was used to train the agent, while the time period between 1 Jan 2017 to 4 Dec 2018 was used to test the agent. We initialize our portfolio with $\$500,000$ in cash. We implement a CRP benchmark where funds are equally distributed among all assets, including the cash asset. We also compare the previous work \cite{jiang2017deep} as a baseline which can be considered as a vanilla model-free DDPG algorithm (i.e., without data augmentation, infused prediction and behavioral cloning)\footnote{In fact, our baseline is superior to the prior work \cite{jiang2017deep} due to the addition of  prioritized experience replay and parameter noise for better exploration.}. It should be noted that, compared to the implementation in \cite{jiang2017deep} our baseline agent is situated in a more realistic trading environment that does not assume an immediate trade execution. In particular, our backtesting engine executes market orders at the open of the next OHLC bar\footnote{Refer to Section \ref{sec3} for details on these terms.}, as well as adds slippage to the trading costs. Additional practical constraints are applied such that fractional trades of an asset are not allowed.

As shown in Table \ref{performance}, adding both DAM and BCM to the baseline leads to a very small increase in Sharpe ratio (from $0.56\%$ to $0.57\%$) and Sortino ratio (from $0.78\%$ to $0.79\%$). There are two hypotheses that we can draw here. First, we hypothesize that the vanilla DDPG algorithm (in its usage within our framework) is not over-fitting to the data since the use of data augmentation has a relatively small impact on the performance. Similarly, it could be hypothesized that the baseline is efficient in its use of available signals in the training set, such that, the addition of behavior cloning by itself, has a negligible improvement of performance. 
By making use of just IPM, we observe a significant improvement over the baseline in terms of Sharpe (from $0.56$ to $0.61$) and Sortino (from $0.78$ to $0.87$) ratios, indicating that we are able to get more returns per unit risk taken. However, we note that the volatility of the portfolio is significantly increased (from $12.79\%$ to $14.14\%$). This could be due to the agent over-fitting to the training set, with the addition of this module. Therefore, when the agent sees the testing set, its actions could result in more frequent losses, thereby increasing volatility. 

To reduce over-fitting, we integrate IPM with DAM. As can be seen in Table \ref{performance}, the Sharpe ratio is actually reduced slightly compared to the IPM case (from $0.61$ to $0.59$) although volatility is significantly reduced (from $14.14\%$ to $12.84\%$). This reduces the overall risk of the portfolio. Moreover, this substantiates our hypothesis of the agent over-fitting to the training set. One possible reason for IPM over-fitting to the training set while the baseline agent does not is the following: the IPM is inherently more complex due to the presence of a prediction model. The increased complexity in terms of larger network structure essentially results in higher probability of over-fitting to the training data. As such, using DAM to diversify the training set is highly beneficial in containing such model over-fitting.

One drawback of using IPM+DAM is the decreased return to risk ratios as mentioned above. To mitigate this, we make use of all our contributions as a single framework (i.e. IPM+DAM+BCM). As shown in Table \ref{performance}, we are not only able to recover the performance of the original model trained using just IPM, but surpass the IPM performance in terms of both Sharpe (from $0.61$ to $0.63$) and Sortino ratios (from $0.87$ to $0.89$). It is worthwhile to note that addition of BCM has a strong impact compared to IPM+DAM in reducing volatility (from $12.84\%$ to $12.77\%$) and MDD (from $12.7\%$ to $12.4\%$). 
The observation of behavior cloning reducing downside risk is further substantiated when comparing the model trained via IPM versus the model trained with IPM and BCM; the Sharpe and Sortino ratios are maintained though annualized volatility is significantly reduced. In addition, when comparing baseline versus baseline with BCM, we see that MDD is reduced and Sharpe and Sortino ratios have small improvements.

We can conclude that IPM significantly improves portfolio management performances in terms of Sharpe and Sortino ratios. This is particularly attractive for investors who aim to maximize their returns per unit risk. In addition, DAM helps to prevent over-fitting especially in the case of larger and more complex network architectures. However, it is important to note that this may impact risk to reward performance. BCM as envisioned, helps to reduce portfolio risk as seen by its ability to either reduce volatility or MDD across all runs. It is also interesting to note its ability to slightly improve the risk to return (reward) ratios. 
Enabling all three modules, our proposed model-based approach can achieve significant performance improvement as compared with benchmark and baseline. In addition, we provide details on an additional risk adjustment module (that can adjust the reward function contingent on risk) with experimental results in supplementary material. Usage of the presented modules with a PPO based on-policy algorithm is also provided.

\section{Conclusion}\label{sec5}
In this paper, we proposed a model-based deep reinforcement learning architecture to solve the dynamic portfolio optimization problem. To achieve a profitable and risk-sensitive portfolio, we developed infused prediction, GAN based data augmentation and behavior cloning modules, and further integrated them into an automatic trading system. The stability and profitability of our proposed model-based RL trading framework were empirically validated on several independent experiments with real market data and practical constraints. Our approach is applicable not only to the financial domain, but also to general reinforcement learning domains which require practical considerations on decision making risk and training data scarcity. Future work could test our model's performance on gaming and robotics applications.



\bibliography{example_paper}
\bibliographystyle{icml2019}

\appendix
~\newpage~\newpage

\section{Algorithms}
Our proposed model-based off-policy actor-critic style RL architecture is summarized in Algorithm \ref{algo1}. 
\begin{breakablealgorithm}
    \caption{Dynamic portfolio optimization algorithm (off-policy version).}
	\label{algo1}
    \small
	\begin{algorithmic}[1]
		\STATE {\bfseries Input:} Critic $Q(\tilde{s}, a | \theta^Q)$, actor
		$\mu(\tilde{s} | \theta^{\mu})$ and {perturbed} actor networks $\mu(\tilde{s} | \theta^{\tilde{\mu}})$ with weights $\theta^{Q}$, $\theta^{\mu}$,  $\theta^{\tilde{\mu}}$ and standard deviation of parameter noise $\sigma$.
		\STATE Initialize target networks $Q'$, $\mu'$ with weights $\theta^{Q'}
		\leftarrow \theta^{Q}$, $\theta^{\mu'} \leftarrow \theta^{\mu}$ 
		\STATE Initialize replay buffer $R$
		\FOR{episode $= 1, \dots, M$}
		\STATE Receive initial observation state $s_1$
		\FOR{$t = 1, \dots, T$}
		\STATE {Predict future price tensor $\mathbf{x}_{t+1}$ with {prediction models} using $s_t$} and form augmented state $\tilde{s}_t$
		\STATE Use perturbed weight $\theta^{\tilde{\mu}}$ to select action ${a}_t = \mu(\tilde{s}_t | \theta^{\tilde{\mu}}) $
		\STATE Execute action $a_t$, observe
		reward $r_t$ and new state $s_{t+1}$
		\STATE {Predict next future price tensor $\mathbf{x}_{t+2}$ using {prediction models} with inputs $s_{t+1}$} and form the augmented state $\tilde{s}_{t+1}$
		\STATE Solve the optimization problem \eqref{expert} for the expert greedy action $\bar{a}_t$ 
		\STATE Store transition $(\tilde{s}_t, a_t,
		r_t, \tilde{s}_{t+1}, \bar{a}_t)$ in $R$
		\STATE Sample a minibatch of $N$ transitions, 
		$(\tilde{s}_i, a_i,
		r_i, \tilde{s}_{i+1}, \bar{a}_i)$, from $R$ via {prioritized replay}, according to temporal difference error 
		\STATE Compute $
		y_i = r_i + \gamma Q'(\tilde{s}_{i+1},
		\mu'(\tilde{s}_{i+1} | \theta^{\mu'}) | \theta^{Q'}) $
		\STATE Update the critic $\theta^Q$ by annealing the prioritized replay bias while minimizing the loss:
		$$\frac{1}{N} \sum_{i=1}^N (y_i -
		Q(\tilde{s}_i, a_i | \theta^Q))^2$$
		\STATE Maintain the ratio between the actual learning rates of the actor and critic
		\STATE Update $\theta^{\mu}$ using the sampled policy gradient:
		\begin{equation*}\label{sec3.1_policy_gradient}
        \begin{aligned}
		\frac{1}{N} \sum_i
		\nabla_{a} Q(\tilde{s}, a | \theta^Q)|_{\tilde{s} = \tilde{s}_i, a = \mu(\tilde{s}_i|\theta^{\mu})}\nabla_{\theta^\mu} \mu(\tilde{s} | \theta^\mu)|_{\tilde{s} =\tilde{s}_i}
        \end{aligned}
        \end{equation*}
        \STATE Calculate the expert auxiliary loss $\bar{L}$ in \eqref{expert_loss} and update $\theta^\mu$ using $\nabla_{\theta^\mu}\bar{L}^\mu$ with factor $\lambda$ 
		\STATE Update the target networks:
		$
		\theta^{Q'} \leftarrow \tau \theta^{Q} + (1 - \tau) \theta^{Q'},
		$
		$
		\theta^{\mu'} \leftarrow \tau \theta^{\mu} +
		(1 - \tau) \theta^{\mu'}
		$ 

		\STATE Create adaptive actor weights $\tilde{\mu}'$ from current actor weight $\theta^{\mu}$ and current $\sigma$: $\theta^{\tilde{\mu}'} \leftarrow \theta^{\mu} + \mathcal{N}(0, \sigma)$ 
		\STATE Generate adaptive perturbed actions $\tilde{a}'$ for the sampled transition starting states $\tilde{s_i}$: 
		$\tilde{a}'=\mu(\tilde{s_i}|\theta^{\tilde{\mu}'})$. With previously calculated actual actions $a = \mu(\tilde{s}_i|\theta^{\mu})$, calculate the mean induced action noise: 
		$
		d(\theta^{\mu}, \theta^{\tilde{\mu}'})=\sqrt{\frac{1}{N}\sum^{N}_{i=1} \mathbb{E}_s \left[ (a_i-a'_i)^2 \right] } 
		$
		\STATE Update $\sigma$: if $d(\theta^{\mu}, \theta^{\tilde{\mu}'})\leq \delta$, $\sigma \leftarrow \alpha\sigma$, otherwise $\sigma\leftarrow\sigma/\alpha$

		\ENDFOR
		\STATE Update perturbed actor: $\theta^{\tilde{\mu}} \leftarrow \theta^{\mu} + \mathcal{N}(0,\sigma)$
		\ENDFOR
	\end{algorithmic}
\end{breakablealgorithm}

An on-policy version of our proposed model-based PPO style RL architecture can be found in Algorithm \ref{ppo}.

\begin{breakablealgorithm}
	\caption{Dynamic portfolio optimization algorithm (on-policy version).}
	\label{ppo}
    \small
	\begin{algorithmic}[1]
		\STATE {\bfseries Input:} Two-head policy network $\mu$ with weights  $\theta^{\mu}$, policy head $\mu^{\text{policy}}: \widetilde{\mathcal{S}}\rightarrow\mathbb{R}^{2m+2}$ and value head $\mu^{\text{value}}: \widetilde{\mathcal{S}}\rightarrow \mathbb{R}$, and clipping threshold $\epsilon$.
		\FOR{episode $= 1, \dots, M$}
		\STATE Receive initial observation state $s_1$
		\FOR{$t = 1, \dots, T$}
		\STATE {Predict future price tensor $\mathbf{x}_{t+1}$ with {prediction models} using $s_t$} and form augmented state $\tilde{s}_t$
		\STATE Use the policy-head of policy network to produce the action mean $\mathbf{m}(\tilde{s}_t)\in\mathbb{R}^{m+1}$ and variance $\mathbf{\Sigma}(\tilde{s}_t)=\text{diag}(\sigma_1,\sigma_2,\dots,\sigma_{m+1})$
		\STATE Execute action $a_t\sim \mathcal{N}(\mathbf{m}(\tilde{s}_t),\mathbf{\Sigma}(\tilde{s}_t))$, observe
		reward $r_t$ and new state $s_{t+1}$
		\STATE Compute the DSR or D3R $d_t$ based on \eqref{DSR} or \eqref{D3R}
		\STATE {Predict next future price tensor $\mathbf{x}_{t+2}$ using {prediction models} with inputs $s_{t+1}$} and form the augmented state $\tilde{s}_{t+1}$
		\STATE Solve the optimization problem \eqref{expert} for the expert greedy action $\bar{a}_t$ 
		\ENDFOR
		\STATE Compute the probability ratio:
		$$r_t(\theta^\mu)=\frac{\mu^{\text{policy}}_{\theta{^\mu}}(a_t|\tilde{s}_t)}{\mu^{\text{policy}}_{\theta_{\text{old}}^{\mu}}(a_t|\tilde{s}_t)}$$
		\STATE Obtain the value estimate from the value head $V(\tilde{s}_t)=\mu^{\text{value}}(\tilde{s}_t)$ and calculate the advantage estimate $\hat{A}_t=\sum_{t'>t}\gamma^{t'-t}d_{t'}-V(\tilde{s}_t)$
		\FOR{$i=1,\dots,N$}
		\STATE Update the policy network using the gradient of the clipped surrogate objective:
		$$\nabla_{\theta{^\mu}^{\text{policy}}}\sum_{t=1}^T\hat{A}_t\min(r_t(\theta^\mu), \text{clip} (r_t(\theta^\mu), 1-\epsilon, 1+\epsilon))$$
		\STATE Update the policy network using the gradient of value loss:
		$$\nabla_{\theta{^\mu}^{\text{value}}}\sum_{t=1}^T\hat{A}_t^2$$
		\STATE Calculate the expert auxiliary loss $\bar{L}$ in \eqref{expert_loss} and update policy network using $\nabla_{\theta{^\mu}^{\text{policy}}}\bar{L}^\mu$ with factor $\lambda$ 
		\ENDFOR
		\ENDFOR
	\end{algorithmic}
\end{breakablealgorithm}

\section{Experiments for On-policy Algorithm}

As shown in Table \ref{performance_ppo}, the proposed IPM can improve the PPO baseline and baseline+BCM model performance. For brevity, we do not enumerate all module combinations. The purpose is to show the proposed RL architecture could be extended to the on-policy setting.

\begin{figure}[t]
\begin{center}
\centerline{\includegraphics[width=3.5in]{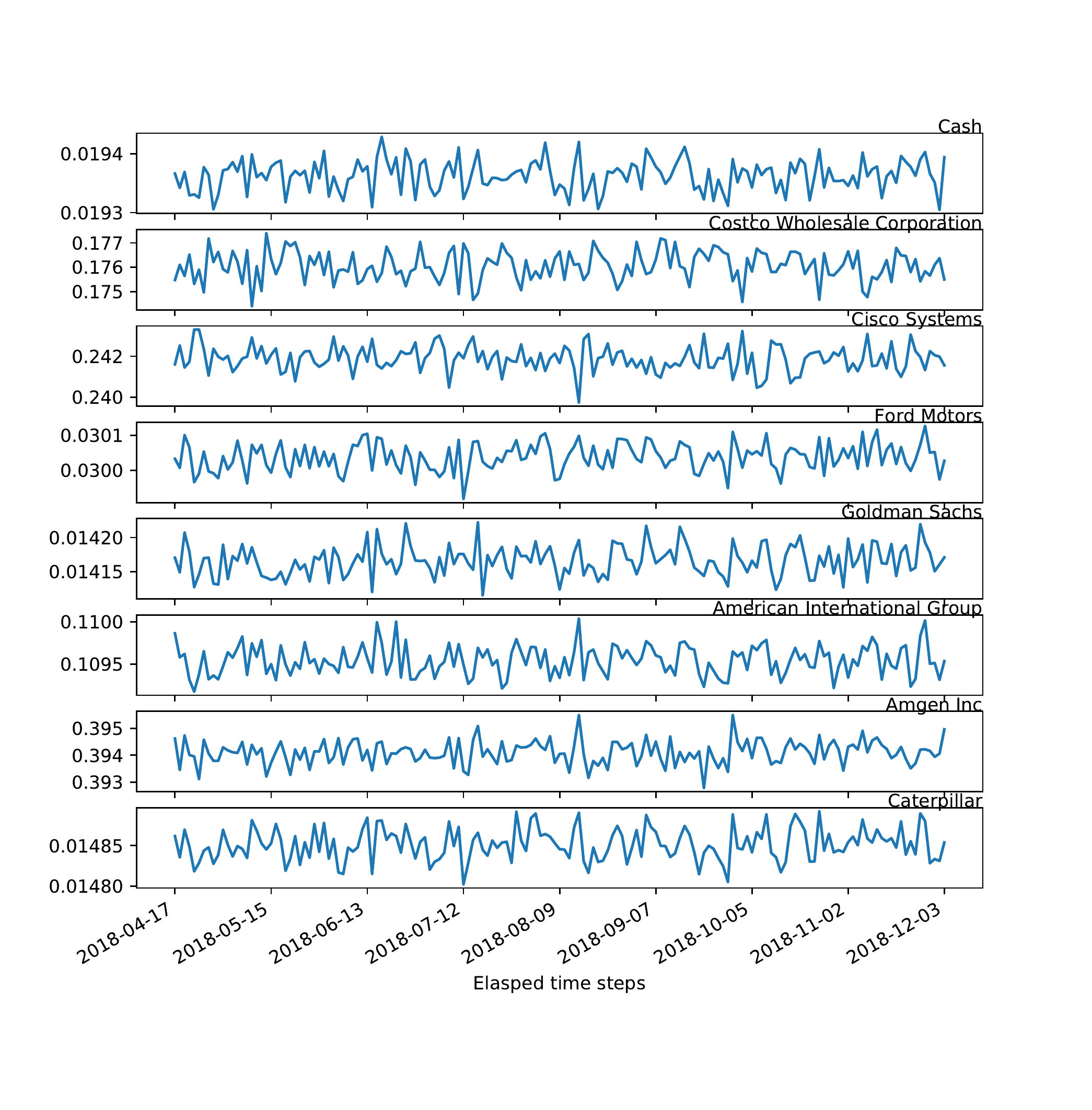}}
\caption{Portfolio weights for different assets (assets from top to bottom are Cash, Costco Wholesale Corporation, Cisco Systems, Ford Motors, Goldman Sachs, American International Group, Amgen Inc and Caterpillar).}
\label{weights}
\end{center}
\vskip -0.2in
\end{figure}

\begin{figure*}[ht]
\begin{center}
\centerline{\includegraphics[height=9in, width=7in]{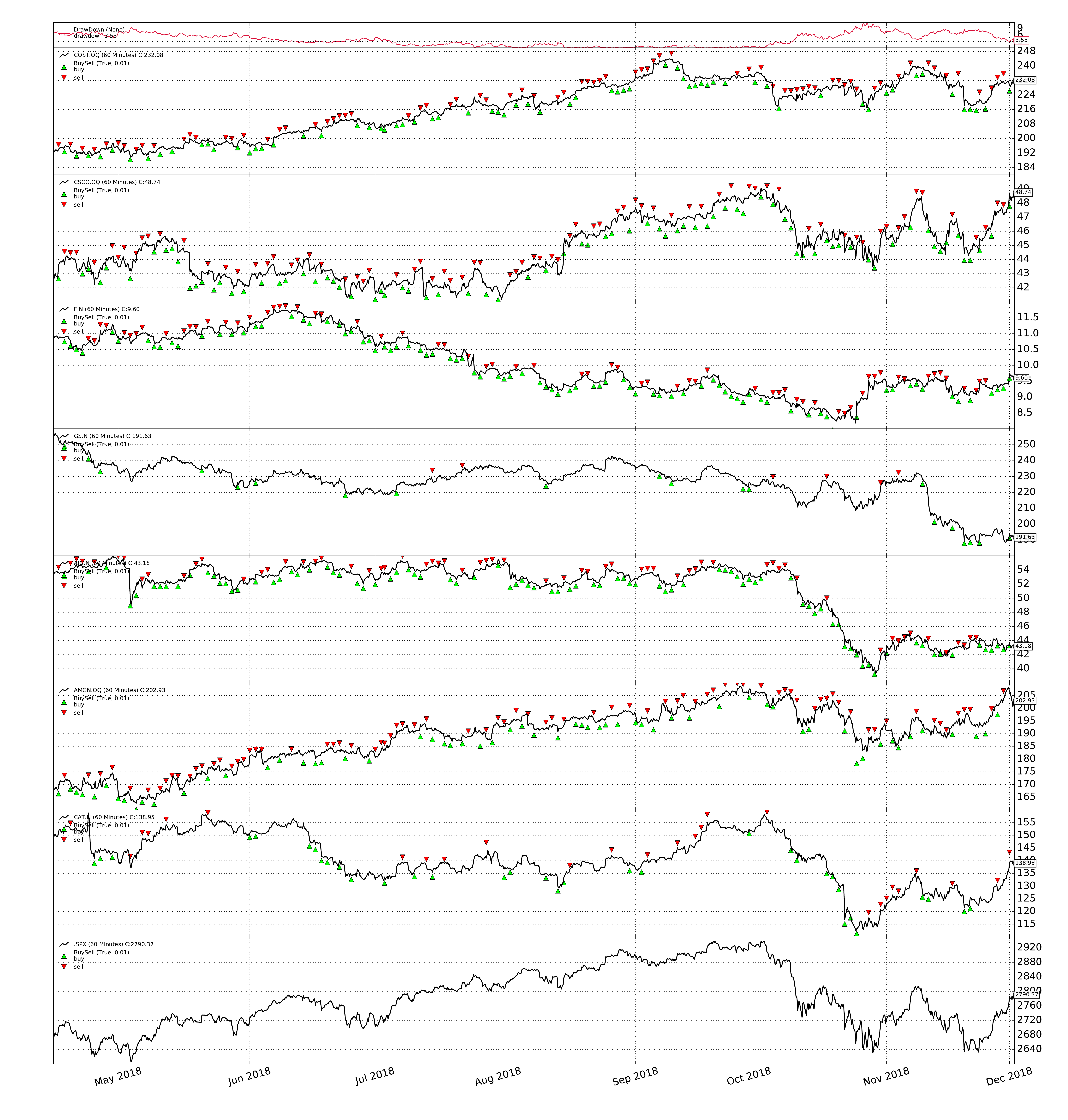}}
\caption{Trading signals (last row is the S\&P 500 index).}
\label{signals}
\end{center}
\end{figure*}

For the trained PPO+BCM model, we arbitrarily chose the testing period between 17 Apr 2018 to 03 Dec 2018 to depict how the portfolio weights evolve over the trading time in Figure \ref{weights}. The corresponding trading signals are shown in Figure \ref{signals}. It can be seen from both figures that the portfolio weights produced by the RL agent can adapt to the market change and evolve as time goes by without falling into a local optimum. 

\section{Hyperparameters for Experiments}
We report the hyperparameters for the actor and critic networks in Table \ref{hyper_ddpg}.

Nonlinear Dynamic Boltzmann machine in the infused prediction module uses the following hyper-parameter settings: delay $d=3$, decay rates $\lambda = [0.1, 0.2, 0.5, 0.8]$ i.e. $k=4$, learning rate was $10^{-3}$, with standard RMSProp optimizer. The input and output dimensions were fixed at three times the number of assets, corresponding to the high, low and close percentage change values. The RNN layer dimension is fixed at 100 units with a $\mathtt{tanh}$ nonlinear activation function. A zero mean, $0.01$ standard deviation noise was applied to each of the input dimensions at each time step, in order to slightly perturb the inputs to the network. This injected noise was cancelled by applying a standard\footnote{As implemented in scientific computing package Scipy.} Savitzky-Golay (savgol) filter with window length $5$ and polynomial order $3$.

For the variant of WaveNet \cite{pmlr-v80-oord18a} in the infused prediction module, we choose the number of dilation levels $L$ to be 6, filter length $f$ to be $2$, the number of filters to be 32 and the learning rate to be $10^{-4}$. Inputs are scaled with min-max scaler with a window size equal to the receptive field of the network, which is $f^L+\sum^{L-2}_{i=0} f^i=95$.

For the DAM, we select the number of training set to be 30,000, noise latent dimension $H$ to be 8, batch size to be 128, time embedding $k_1$ to be 95, generator RNN hidden units to be 32, discriminator RNN hidden unit to be 32 and a learning rate $10^{-3}$. For each episode in Algorithm \ref{algo1}, we generate and append two months of synthetic market data for each asset.

For the BCM, we choose the factor $\lambda=0.1$ to discount the gradient of the log-loss.
\begin{table*}[t]
\caption{Performance for on-policy PPO-style dynamic portfolio optimization algorithm (accnt. and ann. are abbreviations for account and annualized respectively). }
\label{performance_ppo}
\begin{center}
\begin{small}
\begin{tabular}{|c|c|c|c|}
\hline
                      & Baseline & BCM              & IPM+BCM          \\ \hline
Final accnt. value   & 595332   & 672586           & \textbf{711475}  \\ \hline
Ann. return     & 5.23\%   & 15.79\%          & \textbf{18.25\%} \\ \hline
Ann. volatility & 14.24\%  & \textbf{13.96\%} & 14.53\%          \\ \hline
Sharpe ratio          & 0.37     & 1.13             & \textbf{1.26}    \\ \hline
Sortino ratio         & 0.52     & 1.62             & \textbf{1.76}    \\ \hline
$\text{VaR}_{0.95}$                   & 1.41\%   & 1.39\%           & \textbf{1.38\%}  \\ \hline
$\text{CVaR}_{0.95}$                 & 2.10\%   & \textbf{2.09\%}  & 2.20\%           \\ \hline
MDD                   & 22.8\%   & 12.40\%          & \textbf{11.60\%} \\ \hline
\end{tabular}
\end{small}
\end{center}
\end{table*}

\begin{table*}[t]
\caption{Hyperparameters for the DDPG actor and critic networks.}
\label{hyper_ddpg}
\begin{center}
\begin{small}
\begin{tabular}{|c|c|c|}
\hline
                                          & Actor network                                                                                                                      & Critic network         \\ \hline
FE layer type                             & RNN bidirectional LSTM                                                                                                             & RNN bidirectional LSTM \\ \hline
FE layer size                             & 20, 8                                                                                                                              & 20, 8                  \\ \hline
FA layer type                             & Dense                                                                                                                              & Dense                  \\ \hline
FA layer size                             & 256,128,64,32                                                                                                                      & 256,128,64,32          \\ \hline
FA layer activation function              & Leaky relu                                                                                                                         & Leaky relu             \\ \hline
Optimizer                                 & Gradient descent optimizer                                                                                                         & Adam optimizer         \\ \hline
Dropout                                   & 0.5                                                                                                                                & 0.5                    \\ \hline
Learning rate                             & \begin{tabular}[c]{@{}c@{}}Synchronized to be 100 times slower than \\ the critic's actual learning rate \end{tabular} & $10^{-3}$              \\ \hline
Episode length                            & \multicolumn{2}{c|}{650}                                                                                                                                    \\ \hline
Number of episodes                        & \multicolumn{2}{c|}{200}                                                                                                                                    \\ \hline
$\sigma$ for parameter noise              & \multicolumn{2}{c|}{0.01}                                                                                                                                   \\ \hline
Replay buffer size                         & \multicolumn{2}{c|}{1000}                                                                                                                                   \\ \hline
\end{tabular}
\end{small}
\end{center}
\end{table*}

\section{Details of Infused Prediction Module}

\subsection{Nonlinear Dynamic Boltzmann Machine Algorithm}
\begin{figure}[H]
\begin{center}
\centerline{\includegraphics[scale=0.5]{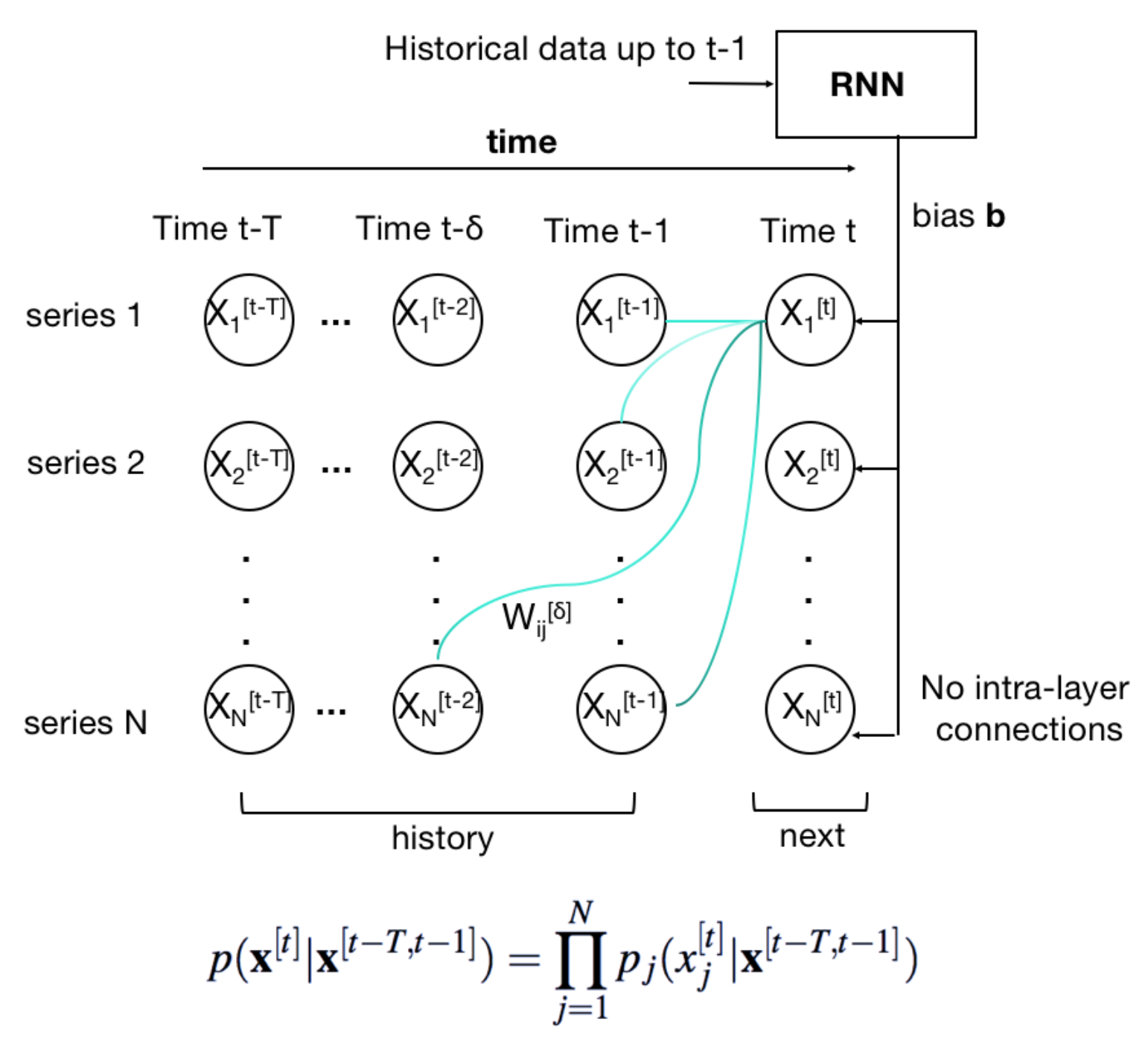}}
\caption{A nonlinear dynamic Boltzmann machine unfolded in time. There are no connections between units $i$
and $j$ within a layer. Each unit is connected to each other unit only in time. The lack of intra-layer connections enables conditional independence as depicted.}
\label{ndybm_fig}
\end{center}
\vskip -0.2in
\end{figure}

In Figure ~\ref{ndybm_fig} we show the typical unfolded structure of a nonlinear dynamic Boltzmann machine. The RNN layer (of a reservoir computing type architecture) \cite{jaeger2003adaptive, jaeger2004harnessing} is used to create nonlinear feature transforms of the historical time-series and update the bias parameter vector as follows:
\begin{equation*}
    \mathbf{b}^{[t]} = \mathbf{b}^{[t-1]} + \mathbf{A}^\top \mathbf{\Psi}^{[t]}
\end{equation*}

Where, $\mathbf{\Psi}^{[t]}$ is a $M \times 1$ dimensional state vector at time $t$ of a $M$ dimensional RNN. $\mathbf{A}$ is the $M \times N$ dimensional learned output weight matrix that connects the RNN state to the bias vector. The RNN state is updated based on the input time-series $\mathbf{x}^{[t]}$ as follows: 
\begin{equation*}
    \mathbf{\Psi}^{[t]} = \mathtt{tanh}\Big (\mathbf{W}_{rnn}\mathbf{\Psi}^{[t-1]} + \mathbf{W}_{in}\mathbf{x}^{[t]} \Big ).
\end{equation*}
Here, $\mathbf{W}_{rnn}$ and $\mathbf{W}_{in}$ are the RNN internal weight matrix and the weight matrix corresponding to the projection of the time-series input to the RNN layer, respectively. 

The NDyBM is trained online to predict $\mathbf{\tilde{x}}^{[t+1]}$ based on the estimated $\mathbf{\mu}^{[t]}$ for all time observations $t=1,2,3, \dots$. The parameters are updated such that the log-likelihood $LL(\mathcal{D}) = \sum_{\mathbf{x}\in \mathcal{D}} \sum_{t} \log p(\mathbf{x}^{[t]}|\mathbf{x}^{[-\infty, t-1]})$, of the given financial time-series data $\mathcal{D}$ is maximized. We can derive exact stochastic gradient update rules for each of the parameters using this objective (can be referenced in the original paper). Such a local update mechanism, allows an $\mathcal{O}(1)$ update of NDyBM parameters. As such a single epoch update of NDyBM occurs in sub-seconds $\ll 1s$ as compared to a tens of seconds update of the WaveNet inspired model. Scaling up to large number of assets in the portfolio, in an online learning scenario this can provide significant computational benefits. 

Our implementation of the NDyBM based infused prediction module is based on the open-source code available at \url{https://github.com/ibm-research-tokyo/dybm}. Algorithm 3 describes the basics steps.

\begin{breakablealgorithm}
    \caption{Online asset price change prediction with the NDybM IPM module.}
    \small
	\begin{algorithmic}[1] \label{alg: ndybm}
    \STATE {\bfseries Require:} All the weight and bias parameters of NDyBM are initialized to zero. The RNN weights, $\mathbf{W}_{rnn}$ initialized randomly from $\mathcal{N}(0,1)$, $\mathbf{W}_{in}$ initialized randomly from $\mathcal{N}(0,0.1)$. The FIFO queue is initialized with $d-1$ zero vectors. $K$ eligibility traces $\boldsymbol{\{\alpha}_k^{[-1]}\}_{k=1}^K$ are initialized with zero vectors 
    \STATE {\bfseries Input:} Close, high and low percentage price change for each asset at each time step
    \FOR{$t \,=\, 0,\,1,\,2,\,...$ \do }
    \STATE Compute $\mathbf{\mu}^{[t]}$ using $\boldsymbol{\mu}^{[t]}= \mathbf{b} + \sum_{\delta=1}^{d-1} \mathbf{F}^{[\delta]}\mathbf{x}^{[t-\delta]} + \sum_{k=1}^K \mathbf{G}_k \, \boldsymbol{\alpha}_k^{[t-1]}$ \& update the bias vector based on RNN layer 
    \STATE Predict the expected price change pattern at time $t$ using $\mathbf{\mu}^{[t]}$
    \STATE Observe the current time series pattern at $\mathbf{x}^{[t]}$
    \STATE Update the parameters of NDyBM based on $\nabla \log p(\mathbf{x}^{[t]}|\mathbf{x}^{[-\infty,t-1]})$
    \STATE Update FIFO queues and eligibility traces by $\alpha_{i,j,k}^{[t]} = \lambda_k \, \alpha_{i,j,k}^{[t-1]} + x_i^{[t-d_{i,j}+1]}$
    \STATE Update RNN layer state vector
    \ENDFOR
    \end{algorithmic}
\end{breakablealgorithm}

\subsection{WaveNet inspired multivariate time-series prediction}
We use an autoregressive generative model, which is a variant of parallel WaveNet \cite{pmlr-v80-oord18a}, to learn the temporal pattern of the percentage price change tensor $\mathbf{x}_t\in\mathbb{R}^{m\times 3}$, which is the part of state space that is assumed to be independent of agent's actions. Our network is inspired by previous works on adapting WaveNet to time series prediction \cite{mittelman2015time, borovykh2017conditional}. We denote the $i^{\text{th}}$ asset's price tensor at time $t$ as $\mathbf{x}_{i,t}=(p_{i,t}, p^{\text{h}}_{i,t}, p^{\text{l}}_{i,t})$.  The joint distribution of price over time $\mathbf{X}=\{\mathbf{x}_1,\dots,\mathbf{x}_T\}$ is modelled as a factorized product of probabilities conditioned on a past window of size $k_1$:
$$
p(\mathbf{X})= \prod^{T}_{t=1} \prod^m_{i=1} p(\mathbf{x}_{i,t} | \mathbf{x}_{t-k_1},\dots,\mathbf{x}_{t-1}, \theta).$$
The model parameter $\theta$ is estimated through maximum likelihood estimation (MLE) respective to $p(\mathbf{X})$. The joint probability is factorized both over time and different assets, and the conditional probabilities is modelled as stacks of dilated causal convolutions. Causal convolution ensures that the output does not depend on future data, which can be implemented by front zero padding convolution output in time dimension such that output has the same size in time dimension as input. 

\begin{figure*}[ht]
\begin{center}
\centerline{\includegraphics[height=8in, width=4in]{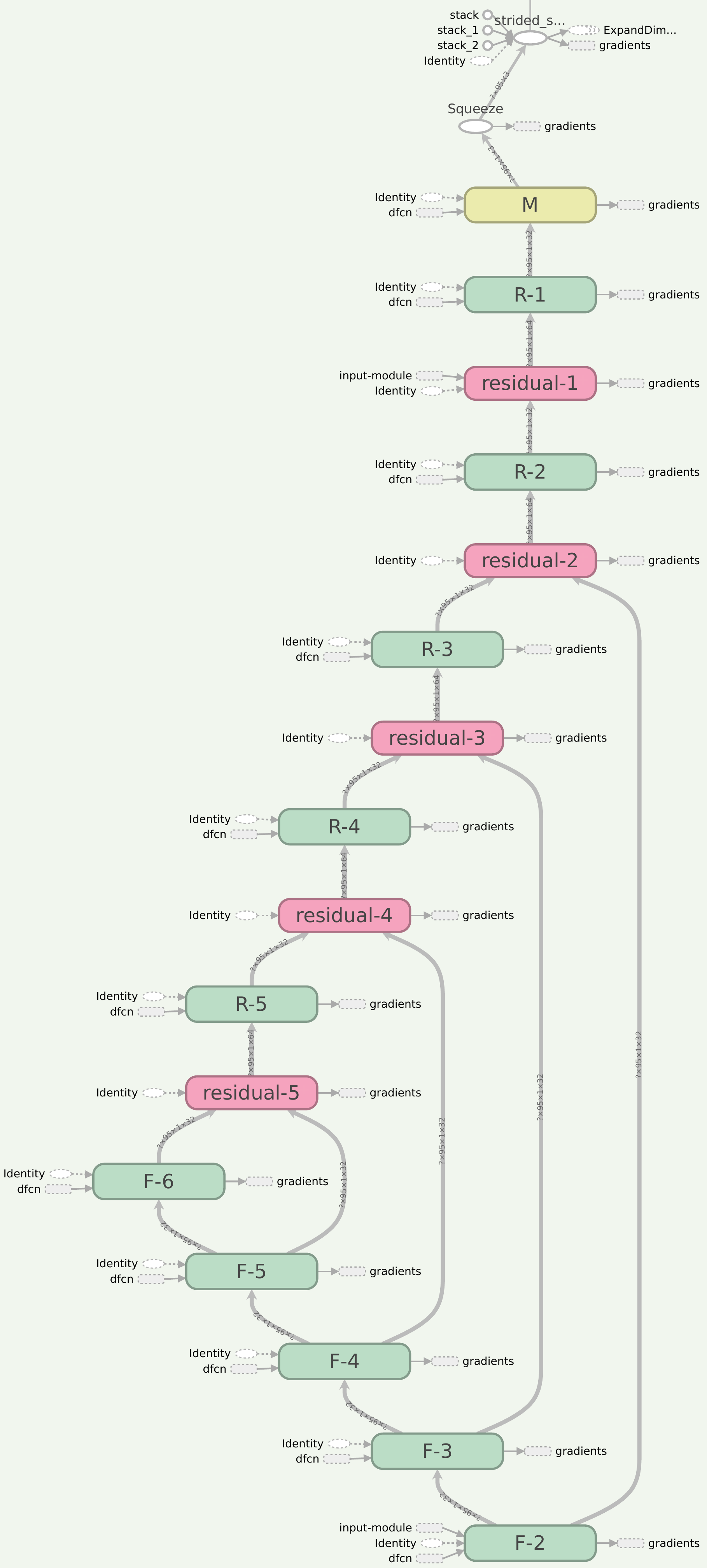}}
\caption{The network structure of WaveNet.}
\label{wavenet}
\end{center}
\end{figure*}

Figure \ref{wavenet} shows the tensorboard visualization of a dilated convolution stack of our WaveNet variant. At each time $t$, the input window $(\mathbf{x}_{t-k_1},\dots,\mathbf{x}_{t-1})\in \mathcal{X}^{k_1}\subset\mathbb{R}^{m\times 3 \times k_1}$ first goes through the common $F_1$ layer, which is a depthwise separable 2d convolution followed by a $1\times 1$ convolution, with time and features (close, high, low) as height and width and assets as channel. Output from $F_1$ is then feed into different stacks of the same architecture as depicted in Figure \ref{wavenet}, one for each different asset. $F_l$ and $R_l$ denotes dilated convolution with filter length $f$ and relu activation at level $l$, which has dilation factor $d_l=f^{l-1}$. Each $R_l$ takes the concatenation of $F_l$ and $R_{l+1}$ as input. This concatenation is represented by the residual blocks in the diagram. The final $M$ layer in Figure \ref{wavenet} is a $1\times 1$ convolution with $3$ filters, one for each of high, low and close, and the output is exactly $\mathbf{x}_{i,t}$. The output from $m$ different stacks is then concatenated to produce the prediction $\mathbf{x}_t$.

The reason for modelling the probabilities as such is two-fold. First, it addresses 
the dependency on historical patterns of the financial market by using a high order autoregressive model to capture long term patterns. Models with recurrent connections can capture long term information since they have internal memory, but they are slow to train. A fully convolutional model can process inputs in parallel, thus resulting in faster training speed compared to recurrent models, but the number of layers needed is linear to $k_1$. This inefficiency can be solved by using stacked dilated convolution. A dilated convolution with dilation factor $d$ uses filters with $d-1$ zero inserted between its values, which allows it to operate in a coarser scale than a normal convolution with same effective filter length. When stacking dilated convolution layers with filter length $f$, if an exponentially increasing dilation factor $d_l=f^{l-1}$ is used in each layer $l$, the effective receptive fields will be $f^L$, where $L$ is the total number of layers. Thus large receptive field can be achieved with having logarithmic many layers to $k_1$, which has much fewer parameters needed compared to a normal fully convolutional model. 

Secondly, by factoring not only over time but over different assets as well, this model makes parallelization easier and potentially has better interpretability. The prediction of each asset is conditioned on all the asset prices in the past window, which makes the model easier to run in parallel since each stack can be placed on different GPUs.

The serial cross correlation $R$ at lag $\ell$ for a pair of discrete time series $x(t), y(t)$ is defined as 
$$ R_{xy}(\ell)=\sum_t x(t)y(t-\ell).$$  

\begin{figure}[H]
\vskip -0.2in
\begin{center}
\centerline{\includegraphics[width=3.3in]{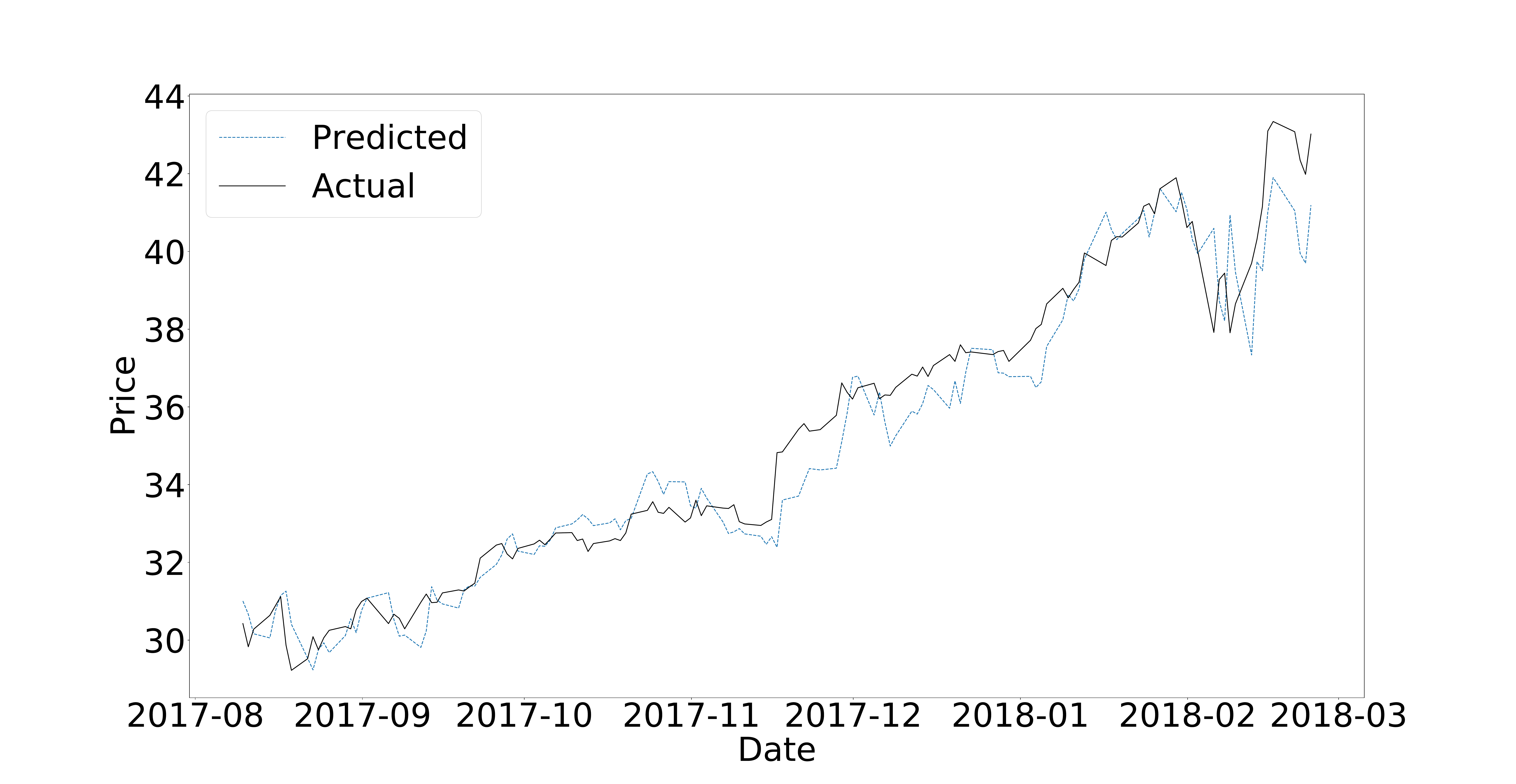}}
\caption{Out of sample actual and predicted closing price movement for asset Cisco Systems.}
\label{tcnn_prediction}
\end{center}
\vskip -0.2in
\end{figure}

\begin{figure}[H]
\begin{center}
\centerline{\includegraphics[width=3.3in]{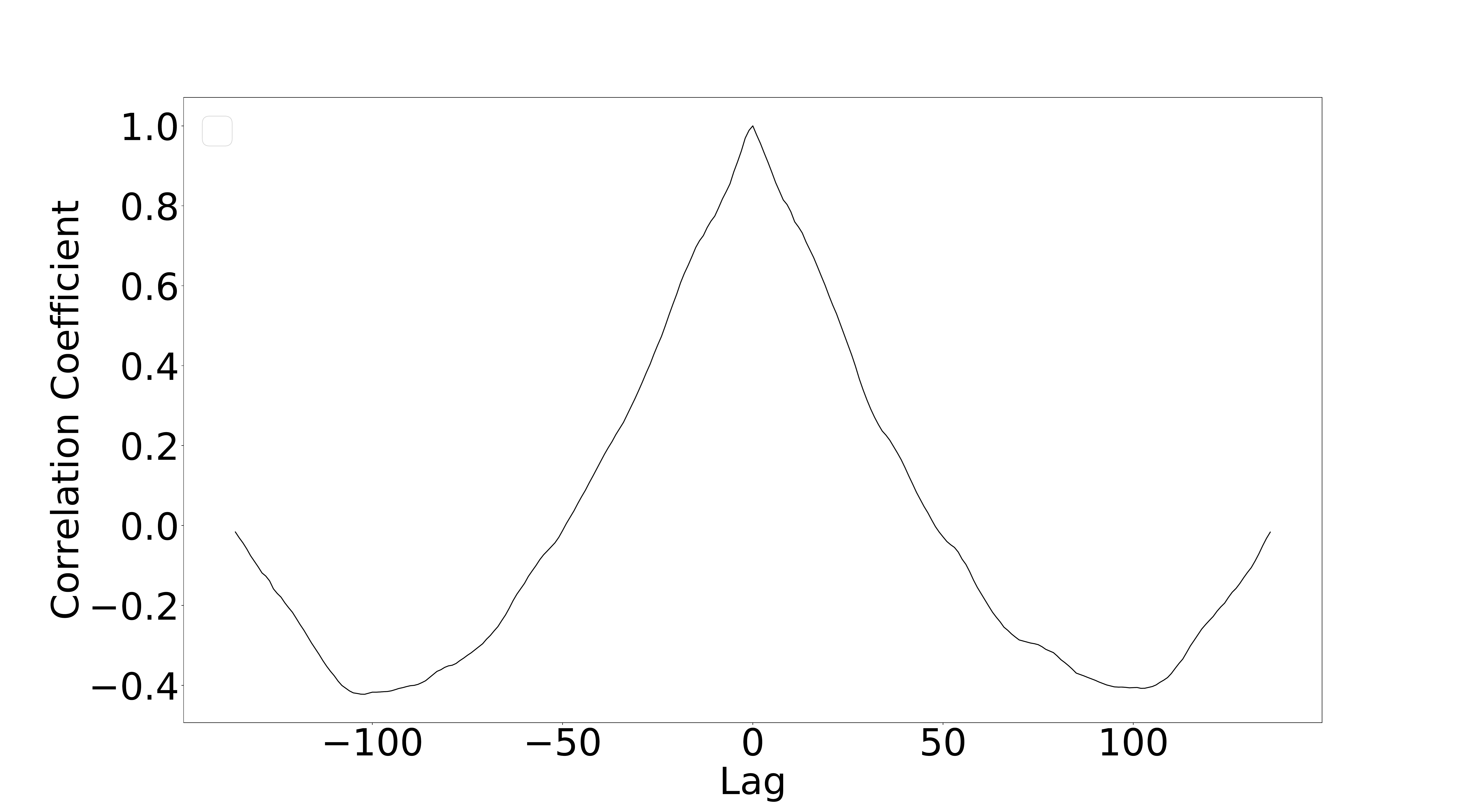}}
\caption{Trend following analysis on predicted closing price movement for asset Cisco Systems.}
\label{tcnn_trend}
\end{center}
\end{figure}

Roughly speaking, $R_{xy}(\ell)$ measure the similarity between $x$ and a lagged version of the $y$. The peak $\alpha = \arg\max_\ell R_{xy}(\ell)$ indicates that $x(t)$ has highest correlation with $y(t+\alpha)$ for all $t$ on average. If $x(t)$ is the predicted time series, and $y(t)$ is the real data, a naive prediction model simply takes the last observation as prediction and would thus have $\alpha=-1$. Sometimes a prediction model will learn this trivial prediction and we call this kind of model trend following. To test whether our model has learned a trend following prediction, we convert the predicted percentage change vector $\mathbf{x}_t$ back to price (see Figure \ref{tcnn_prediction}) and calculated the serial cross-correlation between the predicted series and the actual. Figure \ref{tcnn_trend} clearly shows that there is no trend following behavior as $\alpha=0$.

\section{Details of Data Augmentation Module}
\text{Maximum mean discrepancy (MMD)} \cite{gretton2012} is a pseudometric over $\text{Prob}(\mathcal{X})$, the space of probability measures on some compact metric set $\mathcal{X}$. Given a family of functions $\mathcal{F}$, MMD is defined as 
$$
\text{MMD}(\mathcal{F}, p_a, p_b)=\sup_{f\in\mathcal{F}}\mathbb{E}_{x\sim p_a}[f(x)]-\mathbb{E}_{x\sim p_b}[f(x)].
$$

When $\mathcal{F}$ is the unit ball in a Reproducing Kernel Hilbert Space (RKHS) $\mathcal{H}$ associated with a \textit{universal} kernel $K:\mathcal{X}\times\mathcal{X}\to\mathbb{R}$, i.e. $\{f\in\mathcal{H}:\|f\|_\infty \leq 1\}$, $\text{MMD}(\mathcal{F}, p_a, p_b)$ is not only a pseudometric but a proper metric as well, that is $\text{MMD}(\mathcal{F}, p_a, p_b)=0$ if and only if $p_a=p_b$. One example of a universal kernel is the commonly used Gaussian radial basis function (RBF) kernel $K(x, y)=\exp\left(-\|x-y\|^2/(2\sigma^2)\right)$.

Given samples $\{x_i\}^M_{i=1},\{y_i\}^N_{i=1}$ from $p_a, p_b$, the square of MMD has the following unbiased estimator:
\begin{equation*}
    \begin{aligned}
        \widehat{\text{MMD}}^2=&\frac{1}{ \bigl(\begin{smallmatrix} M \\ 2 \end{smallmatrix}\bigr) }\sum_{i=1}^M\sum_{j\neq i}^M K(x_i,x_j) 
        -\frac{2}{MN}\sum_{i=1}^M\sum_{j=1}^N K(x_i,y_j) \\
        &+\frac{1}{ \bigl(\begin{smallmatrix} N \\ 2 \end{smallmatrix}\bigr) }\sum_{i=1}^N\sum_{j\neq i}^N K(y_i,y_j).
    \end{aligned}
\end{equation*}
We also have the biased estimator where the empirical estimate of feature space means is used
\begin{equation*}
    \begin{aligned}
        \widehat{\text{MMD}}^2_b=&\frac{1}{ M^2 }\sum_{i=1}^M\sum_{j\neq i}^M K(x_i,x_j) 
        -\frac{2}{MN}\sum_{i=1}^M\sum_{j=1}^N K(x_i,y_j) \\
        &+\frac{1}{ N^2 }\sum_{i=1}^N\sum_{j\neq i}^N K(y_i,y_j).
    \end{aligned}
\end{equation*}

Note that the RBF kernel has the following representation
\begin{equation*}
    \begin{aligned}
    K(x,y)=&\exp\left(-\frac{\|x\|^2+\|y\|^2}{2\sigma^2}\right) \exp\left(\frac{\langle x,y \rangle}{\sigma^2}\right) \\
    =&C \sum^\infty_{n=0}\frac{\langle x,y \rangle^n}{\sigma^{2n} n!}\\
    =&C \sum^\infty_{n=0}\frac{\langle\phi_n(x),\phi_n(y)\rangle }{\sigma^{2n} n!}
    \end{aligned}
\end{equation*}
where $C$ is the constant $\exp\left(-||x||^2+||y||^2/(2\sigma^2)\right)$, and $\phi_n(\{\cdot)$ is the feature map of polynomial kernel of degree $n$. Following \cite{yujia2015gmmn}, we can rewrite the \textit{biased} estimator $\widehat{\text{MMD}}^2_b$ with $\phi_n$:
\begin{equation*}
    \begin{aligned}
        \widehat{\text{MMD}}^2_b=& C\sum^\infty_{n=0} \frac{1}{\sigma^{2n} n!} \biggl[ \frac{1}{M^2}\sum_{i=1}^M\sum_{j\neq i}^M \langle\phi_n(x_i),\phi_n(x_j)\rangle \\
        &-\frac{2}{MN}\sum_{i=1}^M\sum_{j=1}^N \langle\phi_n(x_i),\phi_n(y_j)\rangle \\
        &+\frac{1}{ N^2 }\sum_{i=1}^N\sum_{j\neq i}^N \langle\phi_n(y_i),\phi_n(y_j)\rangle \biggr] \\
        =& C\sum^\infty_{n=0} \frac{1}{\sigma^{2n} n!} \left\| \frac{1}{M}\sum^M_{i=1}\phi_n(x_i) - \frac{1}{N}\sum^N_{j=1}\phi_n(y_j) \right\|^2.
    \end{aligned}
\end{equation*}
Therefore minimizing $\widehat{\text{MMD}}^2_b$ can be seen as matching all moments between the two distribution $p_a, p_b$.

Although we can do multivariate two-sample test with MMD, it suffers from the curse of dimensionality. Thus instead of looking at the distribution of percentage change vector $\mathbf{h}_i$, we perform two-sample test on the distribution $p_h$ of percentage change series $h_{i,t}$ itself. Specifically, we use the Kolmogorov-Smirnov test, which is a two-sample test method based on $\|\cdot\|_\infty$ norms between the empirical cumulative distribution functions of the two distributions. For each asset of interest, we divide the data set into a training and a validation set, where the training set is used to train the RGAN model. Then we generate a batch of samples with the trained RGAN. For each generated series, we perform the KS test between it and every series in the validation set and calculates the maximum p-value from which. The average of these maximum p-values, denoted by $\bar{p}_{\max}$, among all the generated samples is then used to determine the goodness of fit of the given generated series. The average $\bar{p}_{\max}$ across all assets in our portfolio is $0.11$, which shows that we cannot reject the null hypothesis that the generated distribution and the data distribution is the same.


\section{Risk-adjustment Module}

In this section, we discuss the risk-adjustment module (RAM) which is aimed to optimize portfolio risk directly. We adapt the framework of recurrent deterministic policy gradient (RDPG, see \cite{heess2015memory}) technique  which is a natural extension of DDPG to memory-based control setting with recurrent neural networks, and adjust our risk preference. 

Note that the sampled minibatch from the replay buffer in Algorithm \ref{algo1} is of the form
$$(s_{t_1},{a}_{t_1},r_{t_1},s_{{t_1}+1}),\dots,(s_{t_K},{a}_{t_K},r_{t_K},s_{{t_K}+1}).$$
The critic and actor networks are then updated according to these {disjointed} (non-successive) samples, i.e., $s_{t_k+1}\neq s_{t_{k+1}}, \forall k={1,\dots,K-1}$. Different from DDPG replay buffer which stores { a time-step experience}, RDPG replay buffer stores { a complete trajectory}. In other words, each sample in the DDPG replay buffer is a state-action-reward-next state tuple, while each sample in the RDPG replay buffer is a trajectory starting from a given initial state. When we sample from the RDPG replay buffer, different trajectories are sampled. 

Different from the original RDPG \cite{heess2015memory}, the update rules of our RDPG are the same as DDPG except that samples from the replay buffer are trajectories instead of disjoint transitions, and that both critic and actor networks in RDPG still take augmented state $\tilde{s}_i$ as inputs. The details of these changes are in Algorithm \ref{ram}


Having access to trajectory sampling also allows us to apply risk-adjustment to our RL objective. Following \cite{moody2001learning}, we consider new risk-aware objectives including differential Sharpe ratio (DSR) and differential downside deviation ratio (D3R). To define the DSR, recall the definition of Sharpe ratio $\text{SR}$ and denote its value at time $t$ by $\text{SR}_t$:
\begin{equation}\label{SR}
    \text{SR}_t = \frac{\mathbb{E}[r_t]}{\sqrt{\text{var}[r_t]}},
\end{equation}
where $r_t$ is the logarithmic rate of return for period $t$. The differential Sharpe ratio $D_t$ is then obtained by considering the moving average of the returns and standard deviation of returns in \eqref{SR}, and expanding to first order in the adaptation rate $\eta$  
\begin{equation}\label{DSR}
    d_t \triangleq \frac{d\text{SR}_t}{d\eta}=\frac{\omega_{t-1}\Delta \nu_t-0.5\nu_{t-1}\Delta \omega_t}{(\omega_{t-1}-\nu^2_{t-1})^{\frac{3}{2}}},
\end{equation}
where $\nu_t$ and $\omega_t$\footnote{This is different from $\mathbf{w}_t$ in the main paper to denote the portfolio weights.} are exponential moving estimates of the first and second moments of $r_t$
\begin{equation*}
\begin{aligned}
    &\nu_t=\nu_{t-1} +\eta\Delta \nu_t=\nu_{t-1} +\eta(r_t-\nu_{t-1}),\\
    &\omega_t=\omega_{t-1} +\eta\Delta \omega_{t}=\omega_{t-1} +\eta(r^2_t-\omega_{t-1}).
\end{aligned}
\end{equation*}
The DSR has several attractive properties including facilitating recursive updating, enabling efficient on-line optimization, weighting recent returns more and providing interpretability \cite{moody2001learning}. 

To define the D3R, we need first to define the downside deviation $dd_T$ as
\begin{equation*}
    dd_T \triangleq \left( \frac{1}{T} \sum_{t=1}^T \min\{r_t, 0\}^2 \right)^{\frac{1}{2}},
\end{equation*}
which is the square root of the average of the square of the negative returns. Using the downside deviation as a measure of risk, we can now define the downside deviation ratio (DDR) as 
\begin{equation}\label{DDR}
    \text{DDR}_T \triangleq \frac{\mathbb{E}[r_t]}{dd_T}.
\end{equation}
The D3R is then defined by considering the exponential moving average of the returns and the squared downside deviation of returns in \eqref{DDR}, and by expanding to the first order in the adaption rate $\eta$ of DDR:
\begin{equation}\label{D3R}
d_t \triangleq \frac{d\text{DDR}_t}{d\eta} = \begin{cases}
\frac{r_t-0.5\nu_{t-1}}{dd_{t-1}} &\text{if $r_t>0,$}\\
\frac{dd_{t-1}^2(r_t-0.5\nu_{t-1})-0.5\nu_{t-1}r_t^2}{dd_{t-1}^3} &\text{otherwise,}
\end{cases}
\end{equation}
where 
\begin{equation*}
\begin{aligned}
    &\nu_t=\nu_{t-1}+\eta(r_t-\nu_{t-1}),\\
    &\text{dd}_t^2=\text{dd}^2_{t-1} + \eta(\min\{r_t,0\}^2-dd^2_{t-1}).
\end{aligned}
\end{equation*}
The D3R rewards the presence of large average positive returns and penalizes risky downside returns. The procedure of obtaining the risk-sensitive RDPG agent for dynamic portfolio optimization is summarized in Algorithm \ref{ram}.

We perform the experiments with initial values $\nu_0=\omega_0=dd_{0}=0$, and add $\varepsilon=10^{-8}$ to the denominators in \eqref{DSR} and \eqref{D3R} to avoid division by zero. The results of DSR can be found in Table \ref{performance_dsr} below. We can observe similar phenomenon to before as shown in Table \ref{performance}, where the rewards were scaled by a fixed number\footnote{All experiment results shown in Table \ref{performance} were generated by scaling the rewards by a factor of $10^3$.}. In particular, the use of IPM drastically improves the Sharpe (from $0.56$ to $0.60$) and Sortino (from $0.78$ to $0.84$) ratios. Furthermore, looking at the model trained with IPM and the model trained with IPM+BCM, we see that the use of BCM is able to slightly improve both Sharpe (from $0.60$ to $0.62$) and Sortino ratios (from $0.84$ to $0.87$). This is consistent with earlier results. Using both IPM+DAM+BCM actually produces the best results compared to just DSR in terms of Sharpe (from $0.56$ to $0.63$) and Sortino (from $0.78$ to $0.88$) ratios. It is worthwhile to note that using IPM+DAM compared to just IPM does not reduce the volatility as in the earlier case, though the Sharpe (from $0.60$ to $0.62$) and Sortino (from $0.84$ to $0.87$) ratios are improved. This could be due to the rewards being risk-adjusted and the DAM allowing better generalization of the model to the testing set.

Results for models with D3R are given in Table \ref{performance_ddr}. We again see that using IPM compared to just D3R improves both Sharpe (from $0.57$ to $0.59$) and Sortino (from $0.79$ to $0.83$) ratios. However, we can observe that the annualized volatility is not significantly impacted when comparing IPM with IPM+DAM+BCM (from $12.75\%$ to $12.74\%$). This could be due to the difference in risk adjustment of the rewards. It is nevertheless important to note that using IPM+DAM+BCM significantly improves Sharpe (from $0.56$ to $0.63$) and Sortino (from $0.78$ to $0.88$) ratios over the model trained with just DSR.

We show that the use of risk-adjusted rewards such as DSR and D3R does not detrimentally impact performance, moreover, it presents an alternative way of scaling the rewards without the need of heuristically setting a scaling factor. 

\begin{table*}[ht]
\caption{Performances for different models (all models are risk-adjusted by DSR risk measure. accnt. and ann. are abbreviations for account and annualized respectively).}
\label{performance_dsr}
\begin{center}
\begin{small}
\begin{tabular}{|c|c|c|c|c|c|c|c|c|}
\hline
                      & DSR  & IPM     & DAM & BCM & IPM+DAM & IPM+BCM & DAM+BCM & IPM+DAM+BCM  \\ \hline
Final accnt. value   & 570909   & 575709  & 570128  & 571595  & 579071      & 578478      & 571880      & \textbf{579444}  \\ \hline
Ann. return     & 7.17\%   & 7.62\%  & 7.10\%  & 7.24\%  & 7.93\%      & 7.88\%      & 7.27\%      & \textbf{7.96\%}  \\ \hline
Ann. volatility & 12.79\%  & 12.75\% & 12.75\% & 12.81\% & 12.75\%     & 12.73\%     & 12.75\%     & \textbf{12.68\%} \\ \hline
Sharpe ratio          & 0.56      & 0.60    & 0.56    & 0.57    & 0.62        & 0.62        & 0.57        & \textbf{0.63}    \\ \hline
Sortino ratio         & 0.78     & 0.84    & 0.78    & 0.79    & 0.87        & 0.87        & 0.79        & \textbf{0.88}    \\ \hline
$\text{VaR}_{0.95}$                   & 1.30\%   & 1.27\%  & 1.28\%  & 1.30\%  & 1.27\%      & 1.26\%      & 1.30\%      & \textbf{1.25\%}  \\ \hline
$\text{CVaR}_{0.95}$                  & 1.93\%   & 1.91\%  & 1.93\%  & 1.94\%  & 1.91\%      & 1.91\%      & 1.93\%      & \textbf{1.90\%}  \\ \hline
MDD                   & 13.70\%  & 12.40\% & 13.60\% & 13.70\% & 12.40\%     & 12.40\%     & 13.70\%     & \textbf{12.20\%} \\ \hline
\end{tabular}
\end{small}
\end{center}
\end{table*}

\begin{table*}[ht]
\caption{Performances for different models (all models are risk-adjusted by D3R risk measure. accnt. and ann. are abbreviations for account and annualized respectively).}
\label{performance_ddr}
\begin{center}
\begin{small}
\begin{tabular}{|c|c|c|c|c|c|c|c|c|}
\hline
                      & DDR     & IPM              & DAM     & BCM     & IPM+DAM          & IPM+BCM          & DAM+BCM & IPM+DAM+BCM      \\ \hline
Final accnt. value   & 571790  & 574791           & 571717  & 571737  & 577833           & 576809           & 571973  & \textbf{578091}  \\ \hline
Ann. return     & 7.26\%  & 7.54\%           & 7.25\%  & 7.25\%  & 7.82\%           & 7.72\%           & 7.27\%  & \textbf{7.84\%}  \\ \hline
Ann. volatility & 12.81\% & 12.75\%          & 12.79\% & 12.79\% & 12.75\%          & 12.75\%          & 12.81\% & \textbf{12.74\%} \\ \hline
Sharpe ratio          & 0.57    & 0.59             & 0.57    & 0.57    & 0.61             & 0.61             & 0.57    & \textbf{0.62}    \\ \hline
Sortino ratio         & 0.79    & 0.83             & 0.79    & 0.79    & \textbf{0.86}    & 0.85             & 0.79    & \textbf{0.86}    \\ \hline
$\text{VaR}_{0.95}$                   & 1.30\%  & \textbf{1.27\%}  & 1.29\%  & 1.30\%  & \textbf{1.27\%}  & \textbf{1.27\%}  & 1.30\%  & \textbf{1.27\%}  \\ \hline
$\text{CVaR}_{0.95}$                  & 1.94\%  & \textbf{1.91\%}  & 1.93\%  & 1.93\%  & \textbf{1.91\%}  & \textbf{1.91\%}  & 1.94\%  & \textbf{1.91\%}  \\ \hline
MDD                   & 13.70\% & \textbf{12.40\%} & 13.60\% & 13.70\% & \textbf{12.40\%} & \textbf{12.40\%} & 13.70\% & \textbf{12.40\%} \\ \hline
\end{tabular}
\end{small}
\end{center}
\end{table*}


\begin{breakablealgorithm}
    \small
    \caption{Risk-adjusted dynamic portfolio optimization algorithm.}\label{ram}
    \begin{algorithmic}[1]
    \STATE {\bfseries Input:} Critic $Q(\tilde{s}, a | \theta^Q)$, actor
	    	$\mu(\tilde{s} | \theta^{\mu})$ and {perturbed} actor networks $\mu(\tilde{s} | \theta^{\tilde{\mu}})$ with weights $\theta^{Q}$, $\theta^{\mu}$,  $\theta^{\tilde{\mu}}$, standard deviation of parameter noise $\sigma$ and risk adaption rate $\eta$.
    \STATE Initialize target networks $Q'$, $\mu'$ 
            with weights $\theta^{Q'}
		    \leftarrow \theta^{Q}$, $\theta^{\mu'} \leftarrow \theta^{\mu}$
    \STATE Initialize replay buffer $R$
    \FOR {episode$=1,\dots,M$}
    \STATE Receive initial observation state $s_1$
    \STATE Initialize $\nu_0$, $\omega_0$ and $dd_0$ for DSR and D3R
    \FOR {$t=1\dots,T$}
    \STATE {Predict future price tensor $\mathbf{x}_{t+1}$ using {prediction              models} with inputs $s_t$} and form augmented state $\tilde{s}_t$
    \STATE Use perturbed weight $\theta^{\tilde{\mu}}$ to select action ${a}_t =         \mu(\tilde{s}_t | \theta^{\tilde{\mu}}) $
    \STATE Take action $a_t$, observe
            reward $r_t$ and new state $s_{t+1}$
    \STATE {Predict next future price tensor $\mathbf{x}_{t+2}$ using {prediction         models} with inputs $s_{t+1}$} and form the augmented state                   $\tilde{s}_{t+1}$
    \STATE Solve the optimization problem \eqref{expert} for the expert greedy           action $\bar{a}_t$ 
    \STATE Compute the DSR or D3R $d_t$ based on \eqref{DSR} or \eqref{D3R}
    \ENDFOR
    \STATE {Store trajectory  $(\tilde{s}_1,{a}_1,d_1, \tilde{s}_2, \bar{a}_t,            \dots, \tilde{s}_T, {a}_T, d_T, \tilde{s}_{T+1}, \bar{a}_T)$ in $R$}
    \STATE Sample a minibatch of size $K$ from replay buffer           $\{(\tilde{s}_{k_1},{a}_{k_1},d_{k_1}, \tilde{s}_{k_2}, \bar{a}_{k_1},  \dots,     \tilde{s}_{k_T}, {a}_{k_T}, d_{k_T}, \tilde{s}_{k_{T+1}},                     \bar{a}_{k_T})\}_{k=1}^K$. 
    \STATE Compute 
           $y_{k_t}=d_{k_t}+\gamma Q'(\tilde{s}_{k_{t+1}},\mu'(\tilde{s}_{k_{t+1}}|\theta^{\mu'})|\theta^{Q'})$
            for all $k=1,\dots, K$ and { $t=1,\dots,T$}
    \STATE Update the critic $\theta^Q$ by minimizing the loss:
            $$\frac{1}{{TK}}{\sum_{t=1}^{T}}\sum_{k=1}^{{K}}(y_{k_t}-Q(\tilde{s}_{k_t},a_{k_t}|\theta^Q))^2$$
    \STATE Update $\theta^{\mu}$ using the sampled policy gradient:
    \begin{equation*}
    \begin{aligned}
        \frac{1}{{TK}}{\sum_{t=1}^{T}}\sum_{k=1}^{K}&\nabla_a Q(\tilde{s}_{t},a)|\theta^Q)|_{\tilde{s}_t=\tilde{s}_{k_t}, a=\mu(\tilde{s}_{k_t}|\theta^{\mu})}\times\\
        &\nabla_{\theta^\mu}\mu(\tilde{s}_{t}|\theta^\mu)|_{\tilde{s}_t=\tilde{s}_{k_t}}
    \end{aligned}
    \end{equation*}		
    \STATE Calculate the expert auxiliary loss $\bar{L}$ in \eqref{expert_loss} and update           $\theta^\mu$ using $\nabla_{\theta^\mu}\bar{L}^\mu$ with factor $\lambda$ 
    \STATE Update the target networks:
	    	$
	    	\theta^{Q'} \leftarrow \tau \theta^{Q} + (1 - \tau) \theta^{Q'},
		    $
		    $
	    	\theta^{\mu'} \leftarrow \tau \theta^{\mu} +
	    	(1 - \tau) \theta^{\mu'}
	    	$ 

		\STATE Create adaptive actor weights $\tilde{\mu}'$ from current actor weight                $\theta^{\mu}$ and current $\sigma$: $\theta^{\tilde{\mu}'} \leftarrow               \theta^{\mu} + \mathcal{N}(0, \sigma)$ 
		\STATE Generate adaptive perturbed actions $\tilde{a}'$ for the sampled transition           starting states $\tilde{s_i}$: 
		        $\tilde{a}'=\mu(\tilde{s_i}|\theta^{\tilde{\mu}'})$. With previously calculated actual actions $a = \mu(\tilde{s}_i|\theta^{\mu})$, calculate the mean induced action noise: 
	        	$
	        	d(\theta^{\mu}, \theta^{\tilde{\mu}'})=\sqrt{\frac{1}{N}\sum^{N}_{i=1} \mathbb{E}_s \left[ (a_i-a'_i)^2 \right] } 
	            $
	     \STATE Update $\sigma$: if $d(\theta^{\mu}, \theta^{\tilde{\mu}'})\leq \delta$,            $\sigma \leftarrow \alpha\sigma$, otherwise $\sigma\leftarrow\sigma/\alpha$
		\STATE Update perturbed actor: $\theta^{\tilde{\mu}} \leftarrow \theta^{\mu} +               \mathcal{N}(0,\sigma)$
\ENDFOR
\end{algorithmic}
\end{breakablealgorithm}

\end{document}